\documentclass{article}

\usepackage{microtype}
\usepackage{graphicx}
\usepackage{subcaption}
\usepackage{booktabs, makecell, multirow, xcolor, colortbl}
\usepackage{hyperref}

\usepackage[preprint]{cb}

\usepackage{amsmath}
\usepackage{amssymb}
\usepackage{mathtools}
\usepackage{amsthm}

\usepackage{bm}
\usepackage{pifont}
\usepackage[capitalize,noabbrev]{cleveref}

\theoremstyle{plain}

\theoremstyle{definition}

\theoremstyle{remark}

\usepackage[textsize=tiny]{todonotes}
\usepackage{microtype}

\usepackage{booktabs}
\usepackage{graphicx}
\usepackage{algorithm}
\usepackage{algorithmic}
\usepackage{enumitem}

\newcommand{\cb}{\textsc{ConceptBank}}

\newcommand{\calD}{\mathcal{D}}

\newcommand{\calS}{\mathcal{S}}

\definecolor{Best}{RGB}{250,210,210}
\definecolor{Second}{RGB}{230,250,230}
\definecolor{highlight}{RGB}{30, 144, 255}

\newcommand{\cmark}{\textcolor{green!50!black}{\ding{51}}}

\newcommand{\best}[1]{\cellcolor{Best}\textbf{#1}}
\newcommand{\second}[1]{\cellcolor{Second}\underline{#1}}
\newcommand{\up}[1]{\textcolor{green!50!black}{\textbf{#1}}}

\usepackage{xspace}
\makeatletter
\DeclareRobustCommand\onedot{\futurelet\@let@token\@onedot}
\def\@onedot{\ifx\@let@token.\else.\null\fi\xspace}

\def\eg{\emph{e.g}\onedot} 
\def\ie{\emph{i.e}\onedot} 
\def\cf{\emph{cf}\onedot} 
 \def\vs{\emph{vs}\onedot}

\makeatother

\begin{document}

\twocolumn[
  \cbtitle{
  Taming SAM3 in the Wild: A Concept Bank for Open-Vocabulary Segmentation
  }

  \cbsetsymbol{equal}{*}

  \begin{cbauthorlist}
    \cbauthor{Gensheng Pei}{skku}
    \cbauthor{Xiruo Jiang}{swjtu}
    \cbauthor{Yazhou Yao}{njust}
    \cbauthor{Xiangbo Shu}{njust}
    \cbauthor{Fumin Shen}{uestc}
    \cbauthor{Byeungwoo Jeon}{skku}
  \end{cbauthorlist}

  \cbaffiliation{skku}{Department of Electrical and Computer Engineering, Sungkyunkwan University}
  \cbaffiliation{swjtu}{School of Computing and Artificial Intelligence, Southwest Jiaotong University}
  \cbaffiliation{njust}{School of Computer Science and Engineering, Nanjing University of Science and Technology}
  \cbaffiliation{uestc}{School of Computer Science and Engineering, University of Electronic Science and Technology of China}

  \vskip 0.3in
]

\printAffiliationsAndNotice{}

\begin{abstract}
The recent introduction of \texttt{SAM3} has revolutionized Open-Vocabulary Segmentation (OVS) through \textit{promptable concept segmentation}, which grounds pixel predictions in flexible concept prompts. However, this reliance on pre-defined concepts makes the model vulnerable: when visual distributions shift (\textit{data drift}) or conditional label distributions evolve (\textit{concept drift}) in the target domain, the alignment between visual evidence and prompts breaks down. In this work, we present \textsc{ConceptBank}, a parameter-free calibration framework to restore this alignment on the fly. Instead of adhering to static prompts, we construct a dataset-specific concept bank from the target statistics. Our approach (\textit{i}) anchors target-domain evidence via class-wise visual prototypes, (\textit{ii}) mines representative supports to suppress outliers under data drift, and (\textit{iii}) fuses candidate concepts to rectify concept drift. We demonstrate that \textsc{ConceptBank} effectively adapts \texttt{SAM3} to distribution drifts, including challenging natural-scene and remote-sensing scenarios, establishing a new baseline for robustness and efficiency in OVS. Code and model are available at \url{https://github.com/pgsmall/ConceptBank}.
\end{abstract}

\section{Introduction}

Open-vocabulary segmentation (OVS)~\cite{ovsp,maskclip,zsseg,ovs_pacl} has fundamentally shifted visual recognition from closed-set taxonomies to language-driven generalization. By leveraging the semantic richness of contrastive vision-language models (VLMs)~\cite{clip,align}, modern OVS approaches allow users to segment arbitrary categories using free-form text. Among these, \texttt{SAM3}~\cite{samv3} stands as a pivotal representative. Through its \textit{promptable concept segmentation} mechanism, \texttt{SAM3} does not merely classify regions; it steers the mask generation process directly via flexible concept prompts, offering a compelling vision of a reusable, \emph{segment-anything} foundation model in the open-vocabulary segmentation regime.

\begin{figure}
    \centering
    \includegraphics[width=1\linewidth]{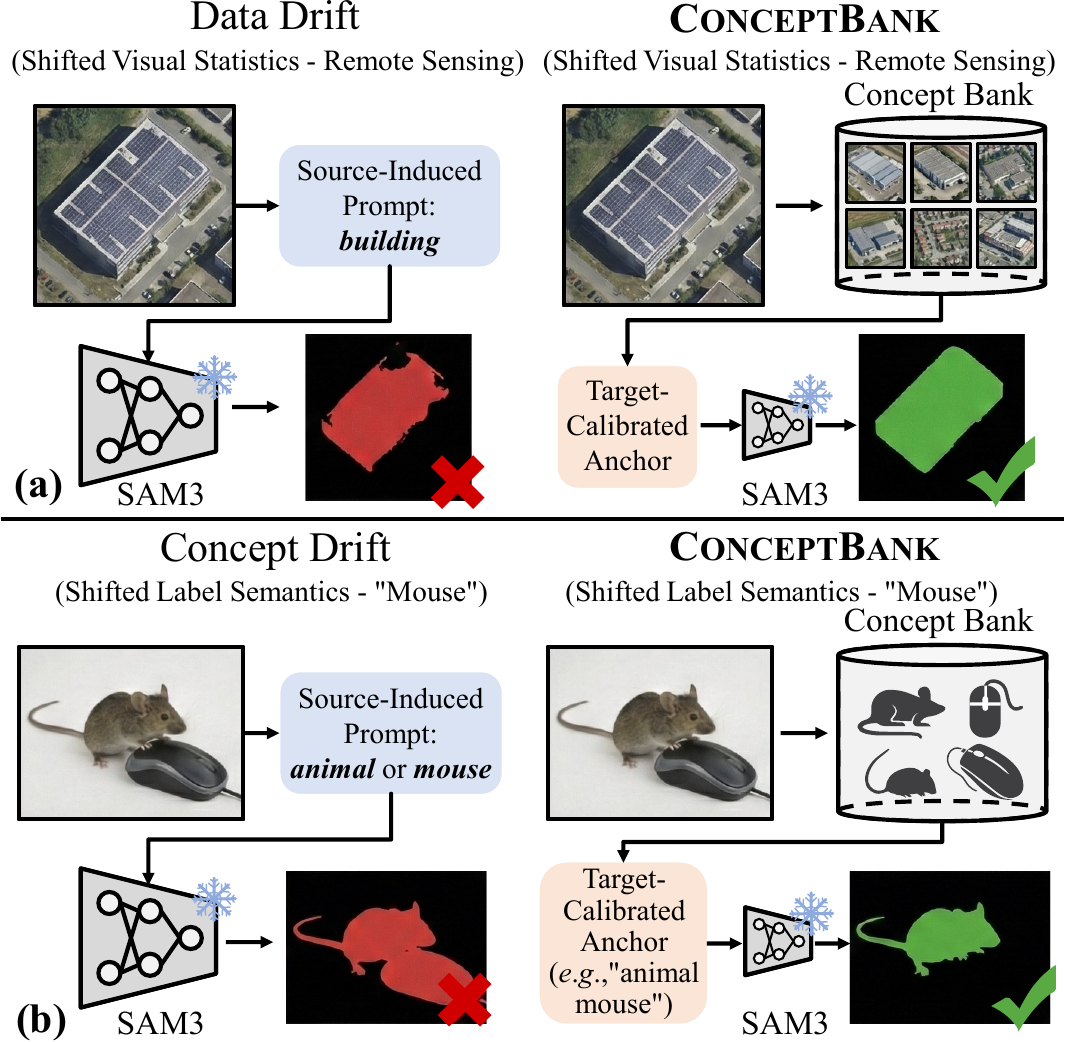}
    \vspace{-0.5cm}
    \caption{\textbf{Data drift and concept drift in \texttt{SAM3}.} We contrast the segmentation outputs of vanilla \texttt{SAM3} and our \cb{} under two failure patterns: (\textbf{a}) \emph{data drift} (shifted visual statistics) and (\textbf{b}) \emph{concept drift} (shifted label semantics). \cb{} restores prompt-mask alignment by replacing static source-induced prompts with target-calibrated anchors from a Concept Bank.}
    \label{fig:drift}
    \vspace{-0.5cm}
\end{figure}

However, the promise of a universal, drop-in segmenter encounters a harsh reality when deployed in the wild: the assumption that a static text embedding (the prompt) remains a valid anchor for visual concepts across diverging domains is often flawed. While \texttt{SAM3} possesses powerful generalization capabilities, its reliance on pre-defined concepts renders it vulnerable to distribution drifts. As illustrated in Fig.~\ref{fig:drift}, this fragility manifests as two distinct yet interconnected failure patterns, as follows:
\begin{itemize}[leftmargin=0.9em, nosep]
    \item \emph{Data Drift} ($P_{\calS}(X) \neq P_{\calD}(X)$): The visual statistics of the target domain $\calD$ often diverge significantly from the source pre-training distribution $\calS$. For instance, in remote sensing or medical imaging, spectral characteristics, bird's-eye viewpoints, and complex background textures perturb the geometry of the dense visual features~\cite{data_drift}. Since \texttt{SAM3} relies on similarity-based matching, geometric distortions cause valid visual evidence to drift away from their corresponding text anchors, leading to segmentation collapse or hallucination.
    \vspace{0.15cm}
    \item \emph{Concept Drift} ($P_{\mathcal{S}}(Y|X) \neq P_{\mathcal{D}}(Y|X)$): The conditional semantics of a label are not immutable; they evolve with context~\cite{concept_drift_1}. A class name such as ``\emph{mouse}'' can denote a computer peripheral in one dataset but refer to a mammal in another domain, leading to incompatible label semantics despite sharing the same name. When a static prompt embedding carries a pre-training-dominant interpretation into a target domain governed by a different labeling rule, the model suffers from semantic misalignment, even if the visual features are robust.
\end{itemize}

Current approaches to mitigate these shifts typically fall into two directions. On one hand, \textit{model fine-tuning} or adapter learning can realign the visual backbone, but this sacrifices the training-free efficiency of foundation models, introduces bias toward the adapted domains, and requires additional optimization and compute~\cite{zsseg,lseg,openseg,sed,cat-seg,segclip,code,talk2dino}. On the other hand, \textit{prompt engineering} attempts to tweak the text side manually, but this is a blind process. It ignores the actual statistical distribution of the target data and often devolves into trial-and-error heuristics. Prompt learning methods improve conditioning when training is allowed~\cite{zhou2022cocoop,khattak2023maple,pratt2023cupl}, yet they still rely on an explicit optimization stage that is misaligned with the plug-and-play deployment goal.
To bridge this gap, we argue for a \textit{data-centric calibration} perspective. Rather than modifying the heavy visual backbone or guessing the optimal text template, we propose to \textbf{\emph{tame}} the model (\ie, \texttt{SAM3}) by recalibrating its concept representations using the target data itself. If the visual evidence shifts, the linguistic anchors must shift to meet it.

In this work, we present \textbf{\cb{}}, a \emph{robust}, \emph{parameter-free} calibration framework for restoring prompt-visual alignment on the fly. Instead of using static, pre-computed text embeddings, \cb{} constructs a dataset-specific dictionary, a concept bank, directly derived from the target domain's statistics (specifically, the support set of $\mathcal{D}$). This bank acts as a dynamic interface between \texttt{SAM3}'s frozen knowledge and the specific nuances of the deployment environment. \cb{} operates via three synergistic mechanisms: (\textit{i}) it estimates class-wise \emph{visual prototypes} to anchor concepts in the target feature space; (\textit{ii}) it applies \emph{representative support mining} to reduce the influence of outliers under severe drift, aligning with robust-estimation principles; and (\textit{iii}) it performs \emph{concept fusion} to synthesize calibrated query embeddings that better match the target labeling semantics.
Crucially, \cb{} requires no gradient updates to \texttt{SAM3}. It functions as a plug-and-play module that strengthens a generic OVS model for distribution-shifted deployment. We demonstrate that this approach is not only conceptually well-motivated but also empirically effective, improving robustness from standard natural-scene benchmarks to remote-sensing imagery.
Our main contributions are summarized as follows:
\begin{itemize}[leftmargin=0.9em, nosep]
\item A drift-centric view of prompt-conditioned OVS is developed by separating \emph{data drift} from \emph{concept drift}, clarifying why static prompts fail under cross-domain use.
\item We propose \cb{}, a \emph{parameter-free} framework tailored for \texttt{SAM3} that efficiently builds a dataset-specific concept bank from target support statistics.
\item We introduce a three-stage construction pipeline comprising \emph{prototype estimation}, \emph{representative support mining}, and \emph{prototype-consistent concept fusion} to stabilize calibration under appearance shifts and semantic relabeling.
\item Extensive experiments on \emph{natural-scene} and \emph{remote-sensing} benchmarks show consistent gains over vanilla prompting and competitive performance against strong baselines in the presence of realistic distribution drift.
\end{itemize}

\begin{figure*}[t]
    \centering
    \includegraphics[width=1.0\linewidth]{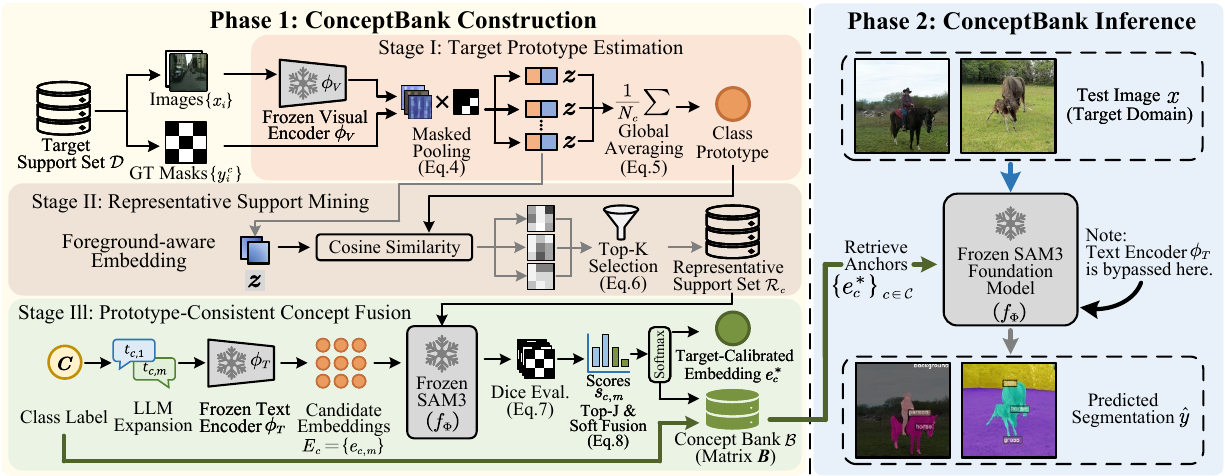}
    \vspace{-0.55cm}
    \caption{\textbf{Illustration of the proposed \cb{} framework.}
    \textbf{Phase 1 (Construction \S\ref{sec:cb_construction})} on target support set $\mathcal{D}$: \textbf{Stage I} estimates class prototypes using frozen $\phi_V$ (masked pooling + class averaging); \textbf{Stage II} mines representative supports $\mathcal{R}_c$ via cosine \texttt{Top}$_K$; \textbf{Stage III} fuses LLM-expanded texts with visual guidance to produce target-calibrated embeddings $e_c^*$, forming Concept Bank $\mathcal{B}$.
    \textbf{Phase 2 (Inference \S\ref{sec:cb_inference})} on target test set: plug $\mathcal{B}$ into frozen \texttt{SAM3} for test segmentation, bypassing the text encoder $\phi_T$.}
    \label{fig:framework}
    \vspace{-0.3cm}
\end{figure*}

\section{Method}
\label{sec:method}

\subsection{Problem Formulation: Distribution Drift in OVS}
Open-vocabulary segmentation is considered on a labeled target-domain dataset $\mathcal{D}=\{(x_i,y_i)\}_{i=1}^{N}$ with $N$ samples, where $x_i\in\mathcal{X}$ denotes an image and $y_i\in\{0,1\}^{H\times W\times|\mathcal{C}|}$ is the corresponding set of category masks at spatial resolution $H\times W$ for $\mathcal{C}$. The category set $\mathcal{C}$ is specified at inference time via language prompts (or concept queries), and the model is expected to generalize beyond the pre-training vocabulary. The samples are assumed to be drawn from an underlying target distribution $P_{\mathcal{D}}$.

A pre-trained and frozen foundation model $\Phi$ (instantiated by \texttt{SAM3}) is adopted, consisting of a visual encoder $\phi_V:\mathcal{X}\rightarrow\mathbb{R}^{d}$ and a text encoder $\phi_T:\mathcal{T}\rightarrow\mathbb{R}^{d}$. Such models follow a \emph{match-and-segment} paradigm: given a prompt $t_c$ (\eg, ``a photo of a [\texttt{class}]''), a query embedding $\bm{e}_c=\phi_T(t_c)$ conditions the mask predictor to produce a binary segmentation $\hat{y}^{c}$. This inference process is summarized as:
\begin{equation}
    \hat{y}^c = f_{\Phi}(x, \bm{e}_c), \quad c \in \mathcal{C}.
    \label{eq:inference_model}
\end{equation}
The key obstacle in transferring $\Phi$ from its pre-training environment $\mathcal{S}$ to the target domain is the misalignment between target visual evidence and source-induced language anchors. This paper characterizes the deployment difficulty as a \emph{dual-drift} phenomenon, including \emph{data drift} ($P_{\mathcal{S}}(X)\neq P_{\mathcal{D}}(X)$) and \emph{concept drift} ($P_{\mathcal{S}}(Y|X)\neq P_{\mathcal{D}}(Y|X)$). Both forms of drift manifest as degraded cross-modal alignment, thereby reducing the reliability of prompt-conditioned masks.

\textbf{\emph{Data Drift}.}
The marginal distribution of visual appearances shifts across domains (\eg, sensor characteristics, spectral responses, viewpoints, illumination, and background context). Even when the semantic definition of class $c$ remains unchanged, the target visual representations $\phi_V(x)$ may move away from the source-induced query embedding $\bm{e}^{\mathcal{S}}_{c}$ in the shared space, leading to weakened alignment:
\begin{equation}
    P_{\mathcal{S}}(X) \neq P_{\mathcal{D}}(X) \Rightarrow \mathbb{E}_{x\sim P_{\mathcal{D}}(X)}[\cos(\phi_V(x), \bm{e}_c^{\mathcal{S}})] \downarrow.
\end{equation}

\textbf{\emph{Concept Drift}.}
The conditional semantics of labels evolve across domains: the visual cues that support label $c$ in $\mathcal{S}$
may differ from those in $\mathcal{D}$.
Consequently, a generic prompt embedding $\bm{e}^{\mathcal{S}}_{c}$ can encode attributes that are irrelevant,
ambiguous, or even contradictory under the target labeling rule, yielding a mis-specified language anchor:
\begin{equation}
    P_{\mathcal{S}}(Y|X) \neq P_{\mathcal{D}}(Y|X) \Rightarrow \bm{e}_c^* \neq \bm{e}_c^{\mathcal{S}}.
\end{equation}

In this work, our goal is to construct a concept bank $\mathcal{B}=\{(c,\bm{e}_c^*)\}_{c\in\mathcal{C}}$ that stores \emph{target-calibrated} query embeddings. Under a fully frozen $\Phi$, the problem reduces to selecting, for each class $c$, a query embedding $\bm{e}_c^*$ that minimizes the target-domain segmentation risk induced by $f_{\Phi}(\cdot,\bm{e}_c)$ on the target-domain dataset $\mathcal{D}$, thereby restoring cross-modal alignment without any parameter update.

\subsection{\cb{}: Construction}\label{sec:cb_construction}

The construction of \cb{} is a one-time, offline process performed on the support set (training split) of the target dataset. It follows a three-stage pipeline that progressively filters noise and realigns the modality gap.

\textbf{\emph{Stage I: Prototype Estimation}.}
To alleviate data drift, we first need to understand where the target visual features lie in the shared embedding space. Relying on global image pooling is insufficient for segmentation tasks as it introduces background noise. Instead, we employ \textit{Mask-Pooled Crop Embeddings}.
For a ground-truth mask $y_i^c$ of class $c$ in image $x_i$, we extract the specific object region $v$. Let $\psi(u; v) \in \mathbb{R}^d$ be the dense visual feature at pixel location $u$ within this crop. We compute the foreground-aware embedding $\bm{z}$:
\begin{equation}
    \bm{z}(v, y_i^c) = \texttt{Norm}\left( \frac{\sum_{u} y_i^c(u) \cdot \psi(u; v)}{\sum_{u} y_i^c(u) + \epsilon} \right),
    \label{eq:mask_pool}
\end{equation}
where $\texttt{Norm}(\cdot)$ denotes $\ell_2$ normalization and $\epsilon$ is a stability constant (set to $10^{-6}$ in this work). We then compute a global class prototype $\bm{p}_c$ by averaging these embeddings across all instances of class $c$ in the support set:
\begin{equation}
    \bm{p}_c = \texttt{Norm}\left( \frac{1}{N_c} \sum_{i, v} \bm{z}(v, y_i^c) \right),
    \label{eq:prototype}
\end{equation}
where $N_c$ denotes the number of available crops (instances) of class $c$ in the support set.
This prototype $\bm{p}_c$ serves as the empirical centroid of the target visual distribution.

\textbf{\emph{Stage II: Representative Support Mining}.}
Prototype estimation in \textbf{\emph{Stage I}} provides a coarse target anchor $\bm{p}_c$. However, directly calibrating language on the full set of crops $\mathcal{V}_c$ is statistically fragile under severe drift.
In practice, $\mathcal{V}_c$ can contain (\textit{i}) long-tail appearances induced by domain shift, (\textit{ii}) heavy occlusions and truncations, (\textit{iii}) background-dominated crops due to imperfect instance extraction, and (\textit{iv}) annotation noise.
Such atypical samples act as high-leverage points in high-dimensional embedding spaces, and can dominate any downstream calibration procedure, yielding a language anchor that explains outliers rather than the target ``\emph{core}''~\cite{huber2011robust,rousseeuw2011robust,hampel1974influence}.
Let $\mathcal{V}_c=\{(v,y)\}$ denote all available crops and their corresponding masks for class $c$ in the support set.
To obtain a stable and domain-representative evidence set, a \emph{representative support set} $\mathcal{R}_c$ is constructed by retaining only crops that are most consistent with the target prototype:
\begin{equation}
    \mathcal{R}_c=\texttt{Top}_{K}\!\left(\mathcal{V}_c;\, (v,y)\mapsto \cos(\bm{z}(v,y),\bm{p}_c)\right),
    \label{eq:support_mining}
\end{equation}
where $\bm{z}(v,y)$ is the mask-pooled crop embedding in Eq.~\eqref{eq:mask_pool}.
Here $\texttt{Top}_{K}(\Omega; g)$ returns the subset of $\Omega$ with the $K$ largest scores under $g$ (see Appendix~\ref{sec:suppA} for the choice of $K$).
This step can be viewed as trimming the tail of the target embedding distribution and preserving a prototype-consistent core.
Consequently, subsequent language calibration is conditioned on visually coherent, frequent target-domain evidence rather than being skewed by outliers under \emph{data drift}.

\textbf{\emph{Stage III: Prototype-Consistent Concept Fusion}.}
Even with representative visual evidence, concept drift implies that the source-induced language anchor $\bm{e}_c^{\mathcal{S}}$ may not describe how class $c$ is realized in the target domain.
A key observation is that this failure mode is \emph{functional}: two prompts with similar embedding similarity can yield markedly different masks.
Therefore, prompt selection is formulated as choosing a query embedding that induces high-fidelity segmentation on the representative supports $\mathcal{R}_c$.

\textbf{Concept Pooling.}
A pool of candidate prompts, $\mathcal{T}_c=\{t_{c,1},\dots,t_{c,M}\}$, is generated by expanding each class name into diverse textual descriptions.
Specifically, an off-the-shelf LLM (\eg, GPT) generates synonyms and attribute-enriched variants conditioned on the dataset documentation (\eg, the original paper and label taxonomy), grounding the candidates in the dataset's label definitions.
Crucially, this step is constrained to text-level rewriting and rephrasing of the provided descriptions, without introducing any new visual concepts, and thus serves purely as non-parametric prompt augmentation.
More details and quality-control protocols are deferred to the Appendix~\ref{sec:suppA}.
Finally, the frozen text encoder maps each candidate to the shared space, yielding text prompt embeddings $\bm{E}_c=\{\phi_T(t)\mid t\in\mathcal{T}_c\}$.

\textbf{Concept Scoring.}
Each candidate embedding $\bm{e}_{c,m}\in\bm{E}_c$ is evaluated by the segmentation quality it induces on $\mathcal{R}_c$.
The score $s_{c,m}$ is defined as the average \texttt{Dice} coefficient over representative supports:
\begin{equation}
    s_{c,m} = \frac{1}{|\mathcal{R}_c|} \sum_{(v, y) \in \mathcal{R}_c} \texttt{Dice}\left( f_{\Phi}(v, \bm{e}_{c,m}), \; y \right).
    \label{eq:scoring}
\end{equation}
Instead of using cosine similarity as a static proxy, we propose a prototype-consistent metric that evaluates candidate concepts via the functional behavior of $f_{\Phi}$ on target data. By grounding the ranking in the model's operational context, our method effectively mitigates \emph{concept drift}.

\textbf{Concept Fusion.}
Relying on a single prompt is often unstable: different phrasings can lead to non-negligible performance variations even when they describe the same concept.
Prompt ensembling has been widely used to mitigate such sensitivity by aggregating multiple semantically compatible prompts, improving robustness and generalization in vision-language transfer.
Accordingly, an index set of high-scoring candidates is selected as
$\mathcal{J}_c=\texttt{Top}_{J}\!\left([M];\, m\mapsto s_{c,m}\right)$, where $[M]=\{1,\dots,M\}$, and Appendix~\ref{sec:suppA} details the choice of $J$.
The final calibrated embedding $\bm{e}_c^*$ is obtained by temperature-scaled soft fusion over $\mathcal{J}_c$:
\begin{equation}
\begin{split}
    \bm{e}_c^* = \sum_{j \in \mathcal{J}_c} w_{c,j} \cdot \texttt{Norm}(\bm{e}_{c,j}), \\
    w_{c,j} = \frac{\exp(s_{c,j}/\tau)}{\sum_{k \in \mathcal{J}_c} \exp(s_{c,k}/\tau)},
\end{split}
\label{eq:fusion}
\end{equation}
where $\tau$ controls the concentration of the mixture weights, serving as a standard smoothing mechanism for softmax-based aggregation~\cite{hinton2015distilling}.
This aggregation yields a calibrated language anchor that remains task-consistent while reducing prompt-level variance, and the resulting pairs $(c,\bm{e}_c^*)$ constitute the concept bank $\mathcal{B}$.

\subsection{\cb{}: Inference}
\label{sec:cb_inference}

Once the concept bank $\mathcal{B}=\{(c,\bm{e}_c^*)\}_{c\in\mathcal{C}}$ is constructed on the target support set, inference in the target domain reduces to prompt-conditioned segmentation with target-calibrated query embeddings.
During inference, the system retrieves $\{\bm{e}_c^*\}_{c\in\mathcal{C}}$ from $\mathcal{B}$ and uses them as language anchors for mask prediction. Each $\bm{e}_c^*$ already aggregates multiple candidate descriptions through \textbf{\emph{Stage III}}, so inference uses one consolidated embedding per class rather than multiple prompts. The text encoder $\phi_T$ is unnecessary in this phase, and $\mathcal{B}$ is stored as a compact embedding matrix $\bm{B}\in\mathbb{R}^{|\mathcal{C}|\times d}$ with one row per class. Given an input image $x$, the inference model outputs masks via:
\begin{equation}
    \hat{y} = f_{\Phi}\!\left(x, \{\bm{e}_c^*\}_{c \in \mathcal{C}}\right).
\end{equation}
\cb{} keeps the match-and-segment paradigm $f_{\Phi}$ intact and only replaces generic source-induced prompts with target-calibrated anchors from $\mathcal{B}$, yielding a plug-in correction for distribution drift without modifying $\Phi$.

In this work, our \cb{} is a \emph{parameter-free} framework that adapts \texttt{SAM3} for OVS: it never updates the parameters of $\Phi$ and constructs $\mathcal{B}$ via forward-only evaluation and non-parametric aggregation on the target support set. During inference, \cb{} remains strictly feed-forward with $\Phi$ fully frozen. Since each $\bm{e}_c^*$ consolidates multiple candidate descriptions offline and $\phi_T$ is bypassed, the runtime cost matches a single-prompt pipeline and improves over multi-prompt ensembling baselines. No gradient-based updates or support samples are used during inference.
The additional cost is storing $\bm{B}$, which scales linearly with $|\mathcal{C}|$.

\begin{table*}[t]
\centering
\setlength{\tabcolsep}{1.0pt}
\caption{\textbf{Quantitative comparison of OVS on natural-scene datasets.} ``Parameter-free'' denotes methods without learnable parameters or gradient-based optimization. Metric values are mIoU (\%). \colorbox{Best}{\best{bold}} and \colorbox{Second}{\second{underlined}} indicate the best and second-best results, respectively.}
\label{tab:ovs-ns}
\vspace{-0.2cm}
\resizebox{\linewidth}{!}{
\begin{tabular}{l c c ccc ccccc c}
\toprule
\multirow{2}{*}{\textbf{Method}} &
\multirow{2}{*}{\textbf{Pub.\,\&\,Year}} &
\multirow{2}{*}{\textbf{Backbone (Size)}} &
\multicolumn{3}{c}{\textbf{\textit{with background}}} &
\multicolumn{5}{c}{\textbf{\textit{without background}}} &
\multirow{2}{*}{\textbf{Avg.}} \\
\cmidrule(lr){4-6}\cmidrule(lr){7-11}
& & & V21 & PC60 & COCO-O & V20 & PC59 & COCO-S & City & ADE & \\
\midrule
\rowcolor{black!5}
\multicolumn{12}{l}{\textbf{\textit{Training-based}}}\\
GroupViT~\cite{groupvit}  & CVPR'22 & GroupViT (ViT-S/16) &
50.4 & 18.7 & 27.5 & 79.7 & 23.4 & 15.3 & 11.1 & 9.2  & 29.4 \\
TCL~\cite{tcl}            & CVPR'23 & CLIP &
51.2 & 24.3 & 30.4 & 77.5 & 30.3 & 19.6 & 23.1 & 14.9 & 33.9 \\
SegCLIP~\cite{segclip}    & ICML'23 & CLIP &
52.6 & 24.7 & 27.5 & - & - & - & - & - & - \\
CoDe~\cite{code}          & CVPR'24 & CLIP &
57.7 & 30.5 & 32.3 & -   & -   & 23.9 & 28.9 & 17.7 & -   \\
SAM-CLIP~\cite{sam-clip}  & CVPR'24 & CLIP + SAM (ViT-B/16) &
60.6 & 29.2 & -  & -   & -   & 31.5  & -  & 17.1 & -   \\
CLIP-DINOiser~\cite{dinoisers} & ECCV'24 & CLIP + DINO (ViT-B/16) &
62.1 & 32.4 & 34.8 & 80.9 & 35.9 & 24.6 & 31.7 & 20.0 & 40.3 \\
Talk2DINO~\cite{talk2dino} & ICCV'25 & CLIP + DINOv2$^\dagger$ (ViT-B/14) &
65.8 & 37.7 & 45.1 & 88.5 & 42.4 & 30.2 & 38.1 & 22.5 & 46.3 \\
\midrule
\rowcolor{black!5}
\multicolumn{12}{l}{\textbf{\textit{Parameter-free}}}\\
CLIP~\cite{clip}         & ICML'21 & CLIP &
18.6 & 7.8  & 6.5  & 49.1 & 11.2 & 7.2  & 6.7  & 3.2  & 13.8 \\
FreeDA~\cite{freeda} & CVPR'24   & CLIP + DINOv2 (ViT-B/14) &
51.8 & 35.3 & 36.3 & 84.3 & 39.7 & 25.7 & 34.1 & 20.8 & 41.0 \\
GEM~\cite{gem}          & CVPR'24 & CLIP &
46.2 & -   & -   & -   & 32.6 & 15.7 & -   & -   & -   \\
CaR~\cite{CaR}          & CVPR'24 & CLIP &
48.6 & 13.6 & 15.4 & 73.7 & 18.4 & -   & -   & 5.4  & -   \\
LaVG~\cite{lavg}            & ECCV'24 & CLIP + DINO (ViT-B/8) &
62.1 & 31.6 & 34.2 & 82.5 & 34.7 & 23.2 & 26.2 & 15.8 & 38.8 \\
ProxyCLIP~\cite{ProxyCLIP} & ECCV'24 & CLIP + DINOv2$^\dagger$ (ViT-B/14) &
58.6 & 33.8 & 37.4 & 83.0 & 37.2 & 25.4 & 33.9 & 19.7 & 41.1 \\
CLIPtrase~\cite{cliptrase} & ECCV'24 & CLIP &
50.9 & 29.9 & 43.6 & 81.0 & 33.8 & 22.8 & 21.3 & 16.4 & 32.7 \\
ClearCLIP~\cite{clearclip} & ECCV'24 & CLIP &
51.8 & 32.6 & 33.0 & 80.9 & 35.9 & 23.9 & 30.0 & 16.7 & 38.1 \\
SCLIP$^\ast$~\cite{sclip}       & ECCV'24 & CLIP &
59.1 & 30.4 & 30.5 & 80.4 & 34.1 & 22.4 & 32.2 & 16.1 & 38.2 \\
NACLIP$^\ast$~\cite{naclip}     & WACV'25 & CLIP &
58.9 & 32.2 & 33.2 & 79.7 & 35.2 & 23.3 & 35.5 & 17.4 & 39.4 \\
LPOSS~\cite{lposs}  & CVPR'25 & CLIP + DINO (ViT-B/16) &
61.1 & 34.6 & 33.4 & 78.8 & 37.8 & 25.9 & 37.3 & 21.8 & 41.3 \\
CASS$^\ast$~\cite{cass}   & CVPR'25 & CLIP + DINO (ViT-B/8) &
65.8 & 36.7 & 37.8 & 87.8 & 40.2 & 26.7 & 39.4 & 20.4 & 44.4 \\
$\mathcal{DIH}$-CLIP~\cite{dih-clip}  & ICCV'25 & CLIP &
64.2 & 36.0 & 37.4 & 84.9 & 39.7 & 26.7 & 40.2 & 19.6 & 43.6 \\
SFP~\cite{sfp}~\cite{sfp}     & ICCV'25 & CLIP &
63.9 & 37.2 & 37.9 & 84.5 & 39.9 & 26.4 & 41.1 & 20.8 & 44.0 \\
CorrCLIP~\cite{corrclip}          & ICCV'25 & CLIP (ViT-L/14) + DINO (ViT-B/8) + SAM2 (Hiera-L) &
76.7 & 44.9 & 49.4 & 91.5 & 50.8 & \second{34.0} & 51.1 & 30.7 & 53.6 \\
Trident~\cite{trident}           & ICCV'25 & CLIP + DINO (ViT-B/16) + SAM (ViT-B/16) &
67.1 & 38.6 & 41.1 & 84.5 & 42.2 & 28.3 & 42.9 & 21.9 & 45.8 \\
ReME~\cite{reme}              & ICCV'25 & CLIP + DINOv2 (ViT-L/14) + SAM (ViT-L/16) &
82.2 & 44.6 & 48.2 & \second{93.2} & \second{53.1} & 33.3 & 59.0 & 28.2 & 55.2 \\
RF-CLIP~\cite{rf-clip}           & AAAI'26 & CLIP &
67.2 & 37.9 & 39.1 & 87.0 & 41.4 & 27.5 & 43.0 & 21.0 & 45.5 \\
\midrule
SAM3$^\ast$~\cite{samv3} & ICLR'26 & SAM3 (PE-L+/14) &
\second{81.9} & \second{46.1} & \second{65.4} & 88.9 & 50.0 & 33.3 & \second{62.3} & \second{31.8} & \second{57.5} \\
\textbf{\cb{} (Ours)} & - & SAM3 (PE-L+/14) &
\best{87.1} & \best{56.5} & \best{67.9} &
\best{97.4} & \best{63.0} & \best{46.4} & \best{75.1} & \best{43.3} & \best{67.1} \\
\bottomrule
\end{tabular}
}
\parbox{\linewidth}{\footnotesize
~\emph{Notes:} Unless otherwise specified, CLIP uses ViT-B/16~\cite{ViT}. ``$\ast$" denotes results reproduced in our work. ``$\dagger$" indicates DINOv2 equipped with registers~\cite{dinov2-reg}. ``-" indicates results are not reported or applicable in the original paper.}
\end{table*}

\section{Experiments}
\label{sec:experiments}

\subsection{Experimental Setup}
\label{sec:setup}

\textbf{Datasets.}
Experiments follow the standard OVS setting on eight natural-scene datasets, including protocols \emph{with} and \emph{without} a background category.
The \emph{with-background} subset includes Pascal VOC 21 (V21, 21 classes)~\cite{pascal_voc}, Pascal Context 60 (PC60, 60 classes)~\cite{pascal_context}, and COCO-Object (COCO-O, 80 object classes from MS-COCO)~\cite{coco}.
The \emph{without-background} subset includes Pascal VOC 20 (V20, 20 classes)~\cite{pascal_voc}, Pascal Context 59 (PC59, 59 classes)~\cite{pascal_context}, COCO-Stuff (COCO-S, 171 classes)~\cite{stuff}, Cityscapes (City, 19 classes)~\cite{Cityscapes}, and ADE20K (ADE, 150 classes)~\cite{ade}.
Unless stated otherwise, results are reported on the official validation splits using the public class-name lists.
To probe distribution drift in OVS beyond natural-scene imagery, experiments additionally include four remote-sensing datasets, each with an explicit background class: LoveDA (7 classes)~\cite{loveda}, Potsdam~\cite{isprs_potsdam} (6 classes), Vaihingen~\cite{isprs_vaihingen} (6 classes), and iSAID (16 classes)~\cite{isaid}.
These aerial benchmarks exhibit pronounced shifts in acquisition conditions and visual statistics (\eg, viewpoint, scale, and sensor characteristics), offering a challenging evaluation setting for distribution drift in OVS.

\vspace{-0.12cm}
\textbf{Implementation and Evaluation Details.}
All experiments use the official \texttt{SAM3}~\cite{samv3} model with the PE-L+~\cite{pe+} backbone and the default configuration, while keeping $\Phi$ fully frozen. Inputs are resized to $1008\times1008$, matching the native \texttt{SAM3} resolution. Unlike CLIP-style OVS pipelines, generic prompt templates (\eg, ``a photo of a [\texttt{class}]'') are not used. Category names serve as the base prompts.
For concept bank construction, we use GPT to generate synonyms and attribute-enriched descriptions for each class conditioned on the dataset's original paper and label definitions. The prompts used for this generation are provided in the Appendix~\ref{sec:suppD}, and we restrict the generation to text-level rewriting without introducing new visual concepts (\cf \S\ref{sec:cb_construction}-\textbf{\emph{Stage III}}).
During inference, each class uses a single calibrated embedding $\bm{e}_c^*$ from $\mathcal{B}$ and bypasses the text encoder $\phi_T$. All results are single-scale, without test-time augmentation, multi-scale or flip evaluation, and without post-processing or refinement (\eg, DenseCRF~\cite{densecrf} or PAMR~\cite{pamr}). All experiments are run on four NVIDIA V100 GPUs (32\,GB). Following prior works~\cite{sclip,naclip,ProxyCLIP}, performance is measured by mean Intersection-over-Union (\textbf{mIoU}).

\begin{table*}[t]
\centering
\setlength{\tabcolsep}{2.3pt}
\caption{\textbf{Quantitative comparison of OVS on remote-sensing datasets.} ``Parameter-free'' denotes methods without learnable parameters or gradient-based optimization. Metric values are mIoU (\%). \colorbox{Best}{\best{bold}} and \colorbox{Second}{\second{underlined}} indicate the best and second-best results, respectively.}
\label{tab:ovs-rs}
\vspace{-0.25cm}
\resizebox{\linewidth}{!}{
\begin{tabular}{l c c c c c c c}
\toprule
\multirow{2}{*}{\textbf{Method}} &
\multirow{2}{*}{\textbf{Pub.\,\&\,Year}} &
\multirow{2}{*}{\textbf{Backbone (Size)}} &
\multicolumn{4}{c}{\textbf{\textit{with background}}} &
\multirow{2}{*}{\textbf{Avg.}} \\
\cmidrule(lr){4-7}
& & & LoveDA & Potsdam & Vaihingen & iSAID & \\
\midrule
\rowcolor{black!5}
\multicolumn{8}{l}{\textbf{\textit{Training-based}}}\\
SAN~\cite{san}  & CVPR'23 & CLIP (ViT-L/14@336px) &
25.3 & 37.3 & 39.2 & 49.6 & 37.9 \\
Cat-Seg~\cite{cat-seg}  & CVPR'24 & CLIP (ViT-L/14@336px) &
28.6 & 35.8 & 42.3 & \second{53.3} & 40.0 \\
SkySense-O~\cite{skysense-o}  & CVPR'25 & CLIP (ViT-L/14@512px) &
\second{38.3} & \second{54.1} & \second{51.6} & 43.9 & \second{47.0} \\
RSKT-Seg~\cite{rskt-seg}  & AAAI'26 & CLIP (ViT-L/14@336px) + RemoteCLIP$^\ddagger$ (ViT-B/16) + DINO (ViT-B/32) &
33.2  & 38.4 &  42.7 & \best{54.3}  & 42.2 \\
\midrule
\rowcolor{black!5}
\multicolumn{8}{l}{\textbf{\textit{Parameter-free}}}\\
CLIP~\cite{clip}         & ICML'21 & CLIP &
12.4 & 15.6 & 10.8 & 7.5 & 11.6 \\
MaskCLIP~\cite{maskclip}   & ECCV'22 & CLIP &
27.8 & 33.9 & 29.9 & 14.5 & 26.5 \\
GEM~\cite{gem}           & CVPR'24 & CLIP&
31.6 & 39.1 & 36.4 & 17.7 & 31.2 \\
ProxyCLIP~\cite{ProxyCLIP} & ECCV'24 & CLIP + DINOv2$^\dagger$ (ViT-B/14) &
34.3 & 49.0 & 47.5 & 21.8 & 38.2 \\
SCLIP$^\ast$~\cite{sclip}       & ECCV'24 & CLIP &
30.4 & 39.6 & 35.9 & 16.1 & 30.5 \\
SegEarth-OV$^\ast$~\cite{segearth}     & CVPR'25 & CLIP &
36.9 & 48.5 & 40.0 & 21.7 & 36.8 \\
CorrCLIP$^\ast$~\cite{corrclip}          & ICCV'25 & CLIP + DINO (ViT-B/8) + SAM2 (Hiera-L) &
36.9 & 51.9 & 47.0 & 25.5 & 40.3 \\
\midrule
SAM3$^\ast$~\cite{samv3} & ICLR'26 & SAM3 (PE-L+/14) &
35.6 & \second{54.1} & 49.1 & 17.7 & 39.1 \\
\textbf{\cb{} (Ours)} & - & SAM3 (PE-L+/14) &
\best{49.4} & \best{60.5} & \best{63.0} & 35.4 & \best{52.1} \\
\bottomrule
\end{tabular}
}
\parbox{\linewidth}{\footnotesize
~\emph{Notes:} Unless otherwise specified, CLIP uses ViT-B/16~\cite{ViT}. ``$\ast$" denotes results reproduced in our work. ``$\dagger$" indicates DINOv2 equipped with registers. ``$\ddagger$" indicates RemoteCLIP~\cite{remoteclip}, a CLIP-based remote sensing foundation model. }
\vspace{-0.35cm}
\end{table*}

\begin{figure*}[t]
    \centering
    \includegraphics[width=1.0\linewidth]{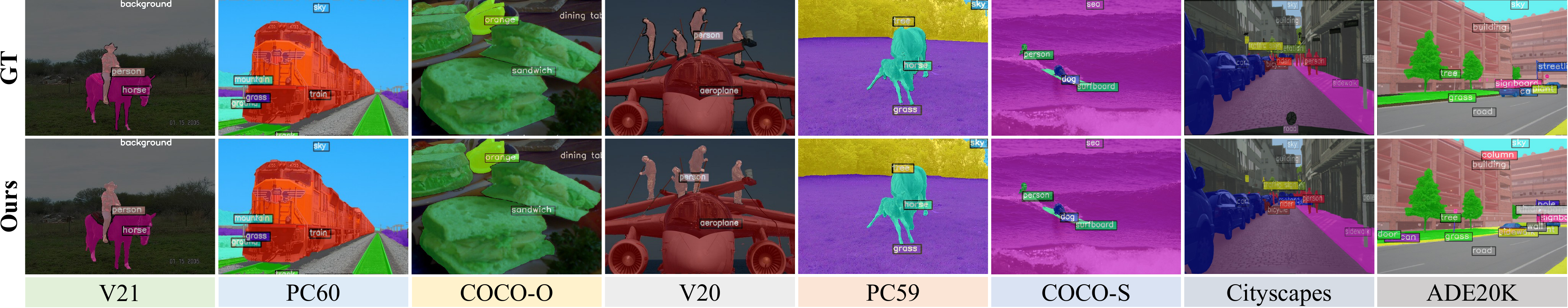}
    \vspace{-0.6cm}
    \caption{\textbf{Qualitative comparison results of open-vocabulary segmentation} on natural-scene datasets. Zoom in for best view.}
    \label{fig:vis_ns}
    \vspace{-0.36cm}
\end{figure*}

\subsection{Comparison with State-of-the-Arts}
\label{sec:sota}

\vspace{-0.1cm}
\textbf{Quantitative Results (Natural-Scene).}
Table~\ref{tab:ovs-ns} reports results on eight natural-scene benchmarks under the OVS protocol.
Most existing methods rely on CLIP-based representations, either through direct prompt matching or by pairing CLIP with auxiliary vision backbones (\eg, DINO/SAM).
\texttt{SAM3} serves as a particularly strong baseline.
Even without calibration, it surpasses all CLIP-based pipelines on average (57.5 mIoU) and rivals the strongest parameter-heavy methods, such as ReME~\cite{reme}, highlighting the advantage of promptable mask decoding backed by a strong visual foundation.
However, \texttt{SAM3} is not fully ``\emph{wild-ready}'': on taxonomy- and context-heavy datasets (\eg, PC60, PC59, and ADE), generic source-induced prompts still misalign with target-domain semantics due to distribution drift, leaving substantial headroom.
\cb{} directly bridges this gap by replacing generic prompts with target-calibrated anchors stored in a concept bank.
Building on the same frozen \texttt{SAM3}, \cb{} improves performance across all datasets, achieving the best overall average of \textbf{67.1} mIoU.
The largest gains appear on PC60 (+\up{10.4}), PC59 (+\up{13.0}), and City (+\up{12.8}), suggesting that these datasets are most susceptible to distribution-drift effects on prompt-conditioned masks.
Since \cb{} introduces no architectural changes or parameter updates to $\Phi$, these improvements are attributed solely to concept-level calibration, effectively taming \texttt{SAM3} for OVS in the wild.

\vspace{-0.09cm}
\textbf{Quantitative Results (Remote-Sensing).}
Table~\ref{tab:ovs-rs} presents results on four remote-sensing benchmarks, where distribution drift is more severe due to acquisition and scale differences.
CLIP-based pipelines remain the dominant paradigm, and even training-based remote-sensing methods or CLIP variants augmented with additional backbones show limited transferability across datasets.
In this regime, \texttt{SAM3} again provides a strong starting point, achieving an average mIoU of 39.1 and already outperforming most CLIP-based baselines. Yet, it still shows clear room for improvement on drift-heavy categories and scenes (notably LoveDA and iSAID).
\cb{} consistently strengthens \texttt{SAM3} under these drifts.
Built on the same frozen backbone and decoder, \cb{} attains the best average of \textbf{52.1} mIoU, improving over \texttt{SAM3} by +\up{13.0} on average.
The performance gains are substantial on all datasets, including LoveDA (+\up{13.8}), Potsdam (+\up{6.4}), Vaihingen (+\up{13.9}), and iSAID (+\up{17.7}). Consequently, our \cb{} achieves the best performance on three out of four benchmarks while remaining clearly ahead on the overall average.
These quantitative results substantiate the indispensability of concept bank calibration for remote-sensing open-vocabulary segmentation, specifically for counteracting semantic misalignment between source prompts and bridging the severe distributional drift from natural pre-training data.

\begin{figure}[t]
    \centering
    \includegraphics[width=1.0\linewidth]{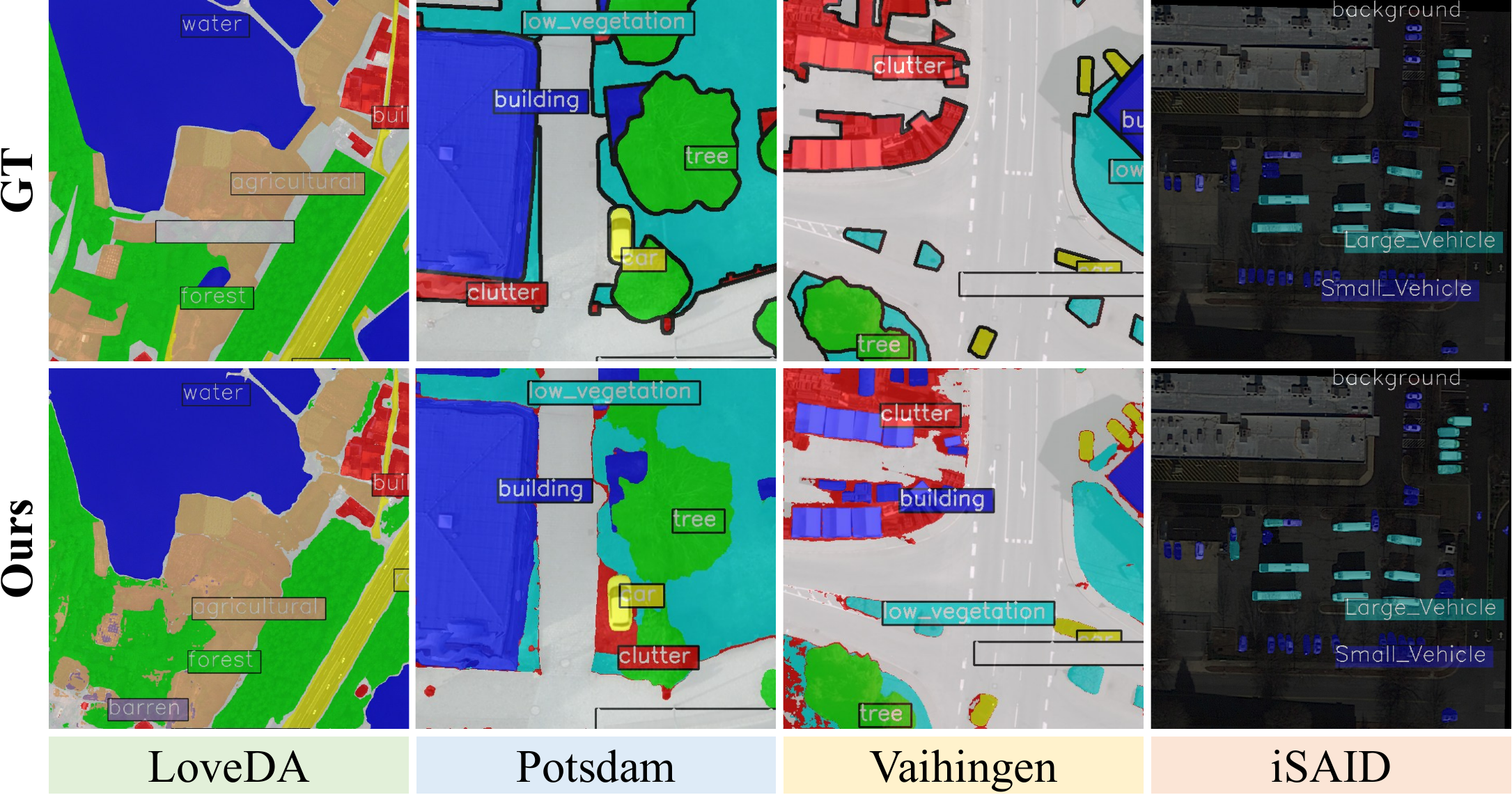}
    \vspace{-0.6cm}
    \caption{\textbf{Qualitative comparison results of open-vocabulary segmentation} on remote-sensing datasets. Zoom in for best view.}
    \label{fig:vis_rs}
    \vspace{-0.25cm}
\end{figure}

\begin{table}[t]
\centering
\setlength{\tabcolsep}{3pt}
\caption{\textbf{Component analysis of \cb{} (\cf \S\ref{sec:cb_construction}).} \textbf{NS} and \textbf{RS} denote Natural-Scene and Remote-Sensing benchmarks.}
\label{tab:ablation-components}
\vspace{-0.25cm}
\resizebox{\linewidth}{!}{
\begin{tabular}{l c c c c c c}
\toprule
\multirow{2}{*}{\textbf{Variant}} & \multicolumn{3}{c}{\textbf{Stage}} & \multicolumn{2}{c}{\textbf{Avg. mIoU}} \\
\cmidrule(lr){5-6}
 & \textbf{I} & \textbf{II} &  \textbf{III} & \textbf{NS} & \textbf{RS} \\
\midrule
\rowcolor{black!5}
Vanilla \texttt{SAM3} (classname only)  &  &  &  & 
57.5 & 39.1 \\
+ Prompt Expansion                      &  &  &  & 
58.0\up{$_{\Delta 0.5}$} & 42.2\up{$_{\Delta 3.1}$} \\
+ Prototype Anchoring                   & \cmark &  &  & 
60.1\up{$_{\Delta 2.6}$} & 45.5\up{$_{\Delta 6.4}$} \\
+ Representative Mining                 & \cmark & \cmark &  & 
62.8\up{$_{\Delta 5.3}$} & 48.2\up{$_{\Delta 9.1}$} \\
\midrule
\rowcolor{green!5}
\textbf{\cb{} (Full)}                   & \cmark & \cmark & \cmark & \textbf{64.7}\up{$_{\Delta 7.2}$} & \textbf{52.1}\up{$_{\Delta 13.0}$} \\
\bottomrule
\end{tabular}
}
\vspace{-0.45cm}
\end{table}

\textbf{Qualitative Results.}
As shown in Fig.~\ref{fig:vis_ns}, \cb{} predictions align closely with GT masks across eight natural-scene benchmarks with diverse taxonomies. The masks essentially preserve scene-level layout and class extents, with most deviations concentrated on fine structures and ambiguous boundaries.
Fig.~\ref{fig:vis_rs} further illustrates \cb{} on four remote-sensing datasets, whose overhead viewpoints and scale patterns differ markedly from natural scenes. Our method still produces coherent large-area land-cover and man-made regions, while remaining errors mostly occur on small objects, thin boundaries, and cluttered areas where visually similar classes are challenging to separate.
Overall, these qualitative results support \cb{} as a concept-level calibration framework that improves prompt-conditioned masks without modifying the frozen \texttt{SAM3}.

\subsection{Ablation Study}
\label{sec:ablation}

This section analyzes the proposed \cb{} to isolate the effect of each design choice.
All ablations follow the protocol in \S\ref{sec:setup} with \texttt{SAM3} kept frozen and identical data splits, and additional ablations are deferred to the \emph{Supplementary Material} (\S\ref{sec:suppA}) due to space constraints.

\textbf{Component Analysis.}
Table~\ref{tab:ablation-components} isolates the contribution of each stage in \cb{} by progressively enabling Stages~I-III.
A plain \texttt{SAM3} baseline already performs strongly on NS (57.5), yet drops on RS (39.1), reflecting the larger distribution drift.
Naive prompt expansion without fusion brings only marginal improvement on NS (+\up{0.5}) but a clearer gain on RS (+\up{3.1}), suggesting that richer descriptions help but do not resolve misalignment by themselves.
Adding Stage~I yields a larger and consistent boost, especially on RS (+\up{6.4}), indicating that grounding concepts in target visual statistics is critical under appearance drift.
Stage~II further improves both settings (NS: +\up{5.3}, RS: +\up{9.1}), supporting its role in filtering atypical or low-quality supports that destabilize calibration.
Finally, Stage~III provides an additional gain and delivers the best overall averages (NS: 64.7, RS: 52.1), showing that consolidating high-scoring candidates into a single calibrated anchor is necessary to fully exploit the expanded concept pool.
Overall, each stage contributes non-trivially, and the full pipeline achieves the most stable improvements across both benchmark families.

\begin{table}[t]
\centering
\setlength{\tabcolsep}{1.pt}
\caption{\textbf{Ablation on prompt expansion strategies.} \textbf{NS} and \textbf{RS} denote Natural-Scene and Remote-Sensing benchmarks.}
\label{tab:ablation-llm}
\vspace{-0.2cm}
\resizebox{\linewidth}{!}{
\begin{tabular}{l c c c}
\toprule
\multirow{2}{*}{\textbf{Variant}} & \textbf{Prompt} & \multicolumn{2}{c}{\textbf{Avg. mIoU}} \\
\cmidrule(lr){3-4}
 & \textbf{Source} & \textbf{NS} & \textbf{RS} \\
\midrule
\rowcolor{black!5}
Vanilla \texttt{SAM3} (classname only)           & - & 57.5 & 39.1 \\
+ Prompt Expansion & Gemini-3 Pro & 57.8\up{$_{\Delta 0.3}$} & 42.1\up{$_{\Delta 3.0}$} \\
\cb{} & Gemini-3 Pro & 64.3\up{$_{\Delta 6.8}$} & 51.8\up{$_{\Delta 12.7}$} \\
\midrule
+ Prompt Expansion & GPT-5.2 & 58.0\up{$_{\Delta 0.5}$} & 42.2\up{$_{\Delta 3.1}$} \\
\rowcolor{green!5}
\textbf{\cb{}} & GPT-5.2 & \textbf{64.7}\up{$_{\Delta 7.2}$} & \textbf{52.1}\up{$_{\Delta 13.0}$} \\
\bottomrule
\end{tabular}
}
\vspace{-0.1cm}
\end{table}

\begin{table}[t]
\centering
\setlength{\tabcolsep}{3.6pt}
\caption{\textbf{Inference efficiency comparison.} Relative FPS measures inference speed normalized to the single-prompt \texttt{SAM3} baseline.}
\label{tab:ablation-speed}
\vspace{-0.2cm}
\resizebox{\linewidth}{!}{
\begin{tabular}{l c c}
\toprule
\textbf{Variant} & \textbf{Text Encoder} & \textbf{Relative FPS} \\
\midrule
\rowcolor{black!5}
Vanilla \texttt{SAM3} (classname only)        & On-the-fly & 1.00$\times$ \\
Vanilla \texttt{SAM3} (prompt expansion)      & On-the-fly & 0.18$\times$ \\
\midrule
\rowcolor{green!5}
\textbf{\cb{} }             & Bypassed   & \textbf{1.25}$\times$ \\
\bottomrule
\end{tabular}
}
\vspace{-0.5cm}
\end{table}

\textbf{Effect of Prompt Expansion and LLM Choice.}
Table~\ref{tab:ablation-llm} reports prompt expansion and then evaluates it under \cb{}.
Prompt expansion alone is modest on Natural-Scene benchmarks (+0.3 with Gemini-3 Pro, +0.5 with GPT-5.2) and more noticeable on Remote-Sensing (+3.0/+3.1), but it remains far from closing the gap induced by drift.
Once the same candidates are calibrated by \cb{}, performance rises sharply to \up{64.3}/\up{51.8} (Gemini-3 Pro) and \up{64.7}/\up{52.1} (GPT-5.2), \ie, +6.8/+12.7 and +7.2/+13.0 over vanilla \texttt{SAM3}.
Equivalently, \cb{} contributes an additional +6.5/+9.7 (Gemini-3 Pro) and +6.7/+9.9 (GPT-5.2) beyond prompt expansion, showing that the gains come from target-conditioned calibration rather than from richer text alone.
The same pattern holds for both LLMs, indicating that our \cb{} is robust to the choice of generator, with GPT-5.2 giving a small but consistent edge.

\textbf{Inference Efficiency.}
Table~\ref{tab:ablation-speed} summarizes runtime throughput \emph{during inference only}.
We report relative FPS for the online segmentation pipeline and exclude Concept Bank construction, which is a one-time offline cost discussed in the Appendix~\ref{sec:suppB}.
Vanilla \texttt{SAM3} with a single class-name prompt sets the reference (1.00$\times$), whereas evaluating multiple prompts per class slows inference sharply (0.18$\times$) due to repeated text encoding and per-prompt mask decoding.
In contrast, \cb{} retrieves pre-calibrated anchors from $\mathcal{B}$ and bypasses the text encoder entirely, so each class uses one consolidated embedding at runtime.
This keeps the online procedure close to the single-prompt baseline and yields a modest speedup in practice (\textbf{1.25}$\times$), while maintaining the same frozen $\Phi$ and decoding path.

\vspace{-0.1cm}
\section{Related Work}
\label{sec:related_work}

\textbf{Open-Vocabulary Segmentation and Foundation Models.}
Semantic segmentation has been historically studied in a closed-set regime with fully supervised architectures such as FCN~\cite{fcn}, DeepLab~\cite{deeplab}, and Transformer-based designs~\cite{xie21_segformer}, evaluated on curated benchmarks (\eg, PASCAL VOC, MS-COCO, ADE20K, Cityscapes, COCO-Stuff). Moving beyond fixed taxonomies motivates open-vocabulary formulations, where early zero-shot transfer leveraged semantic embeddings and class descriptions~\cite{ovsp,zero-shot-semseg}. The modern surge is driven by vision-language models such as CLIP~\cite{clip}, ALIGN~\cite{align}, and EVA~\cite{eva}, together with prompt design/learning strategies that improve text conditioning in downstream tasks~\cite{zhou2022cocoop,khattak2023maple,zhang2022tipadapter,pratt2023cupl}. In parallel, unified decoders and multi-task segmentation frameworks broaden the interface for language-conditioned dense prediction~\cite{x-decoder,zou2023seem,lseg,openseg,glip,regionclip,detic}. Segment-anything style foundation models inject strong mask priors and promptable decoding into segmentation pipelines~\cite{sam,samv2,samv3}, and hybrid systems further couple these priors with VLM embeddings for semantic grounding~\cite{sam-clip}. A recurring bottleneck, however, is that cross-modal alignment learned on pre-training environments can degrade under distribution drift, a phenomenon well documented in concept-drift and data-drift literature~\cite{concept_drift_2,concept_drift_1,data_drift}. This work addresses the same open-vocabulary interface but targets drift explicitly via a parameter-free concept bank that calibrates language anchors to target evidence while keeping $\Phi$ fully frozen.

\textbf{Training-Based and Parameter-Free OVS under Distribution Drift.}
Training-based OVS methods adapt VLMs to dense prediction by learning mask-aware objectives, decoders, or adapters on additional data~\cite{zsseg,ovs_pacl,segclip,cat-seg,groupvit}. Another strong direction fuses complementary foundations, \eg, self-supervised ViTs~\cite{dinov1,dinov2,dinov3,tokencut,uol,cutlearn} and diffusion priors~\cite{diffusion,ovdiff}, to refine localization and objectness, often with distillation-style transfers~\cite{hinton2015distilling,dinoisers,dino.txt,talk2dino}. In contrast, parameter-free OVS keeps backbones frozen and improves inference by reweighting attention, reconstructing patch correlations, suppressing outliers, or enforcing spatial coherence~\cite{maskclip,ProxyCLIP,sclip,naclip,cass,corrclip,kang2025your,reme}. These approaches can be effective but remain sensitive to drift because prompt semantics and visual evidence can become mismatched across domains~\cite{closerlook,concept_drift_2,concept_drift_1,data_drift}. The issue is amplified in remote sensing, where sensor characteristics and viewing geometry induce large appearance shifts. Recent work, therefore, adapts OVS pipelines to aerial benchmarks using either training-based or foundation-model-heavy designs~\cite{remoteclip,segearth,skysense-o,rskt-seg}. Robust estimation principles also emphasize the disproportionate impact of high-leverage outliers in high-dimensional settings, aligning with the need for representative evidence selection before any calibration. In this context, \cb{} departs from heuristic prompt ensembling by constructing a dataset-specific concept bank through prototype anchoring, representative support mining, and task-conditioned concept fusion, thereby strengthening \texttt{SAM3} for open-vocabulary segmentation under realistic distribution drift.

\section{Conclusion}

This work studies why promptable open-vocabulary segmentation with \texttt{SAM3} degrades under distribution drift and presents \cb{}, a parameter-free calibration framework that restores prompt-mask alignment through a dataset-specific concept bank.
\cb{} keeps the match-and-segment operator $f_{\Phi}$ intact and calibrates only the language anchors using target-domain statistics.
It estimates class-wise visual prototypes to anchor target evidence, mines representative supports to reduce the influence of outliers induced by data drift, and fuses candidate concepts to correct mis-specified prompts resulting from concept drift.
Across natural-scene and remote-sensing benchmarks, \cb{} consistently strengthens \texttt{SAM3}'s open-vocabulary segmentation, demonstrating that concept-level calibration is an effective and practical route to taming foundation segmentation models in the wild.

\vspace{0.26cm}
\textbf{Limitation and Future Work.}
Most current OVS methods remain constrained by fixed label taxonomies and static text prompts.
When class definitions are ambiguous, overlapping, or evolve across datasets, the boundary between concept drift and annotation inconsistency becomes ill-posed, and even a calibrated concept bank cannot fully resolve inherently subjective labeling rules.
We aim to generalize this concept-level calibration strategy to other multi-modal foundation models beyond segmentation.

\section*{Impact Statement}

This paper presents work whose goal is to enhance the robustness and adaptability of open-vocabulary segmentation. There are many potential societal consequences of our work, none which we feel must be specifically highlighted here.

\bibliography{ref}
\bibliographystyle{cb}

\newpage
\appendix
\onecolumn

\section*{Appendix}

\section{Additional Ablations}
\label{sec:suppA}

This section reports additional ablations omitted from the main paper due to space constraints.
All settings follow \S\ref{sec:setup} with \texttt{SAM3} fully frozen and identical data splits.
Unless otherwise stated, we use the same benchmark grouping as the main paper:
\textbf{NS} (Natural-Scene) and \textbf{RS} (Remote-Sensing), and report \textbf{Avg. mIoU}.

\vspace{0.05cm}
\subsection{Sensitivity to Representative Mining: \texttt{Top}$_K$}
\label{sec:suppA_topk}

Stage II (\S\ref{sec:cb_construction}) trims atypical supports by retaining the $K$ crops that are most consistent
with the target prototype (Eq.~\eqref{eq:support_mining}).
In our implementation, \texttt{Top}$_K$ corresponds to the per-class budget of representative crops collected in Stage II.
We ablate $K$ over a broad range.
When a class contains fewer than $K$ supports, we use all available crops.

\begin{table}[h]
\centering
\setlength{\tabcolsep}{4pt}
\caption{\textbf{\texttt{Top}-$_K$ sensitivity in Stage II.}
\textbf{NS} and \textbf{RS} denote Natural-Scene and Remote-Sensing benchmarks.}
\label{tab:supp_topk}
\vspace{-0.15cm}
\resizebox{0.46\linewidth}{!}{
\begin{tabular}{c c c >{\columncolor{green!5}}c c c c}
\toprule
\textbf{$K$} & 1 & 5 & 10 & 30 & 50 & 100 \\
\midrule
\textbf{NS Avg. mIoU} & 61.7 & 63.3 & \textbf{64.7} & 64.1 & 64.5 & 64.3 \\
\textbf{RS Avg. mIoU} & 48.1 & 51.0 & 52.1 & 52.3 & \textbf{52.5} & 51.8 \\
\bottomrule
\end{tabular}
}
\vspace{-0.2cm}
\end{table}

\texttt{Top}-$_K$ controls a robustness-coverage trade-off: smaller $K$ yields a tighter ``core'' (stronger trimming),
while larger $K$ increases appearance coverage but risks re-introducing outliers and background-dominated crops.
In practice, performance is typically stable within a mid-range of $K$, motivating a single default ($K=10$) used in the main paper.

\subsection{Sensitivity to Concept Fusion: \texttt{Top}$_J$}
\label{sec:suppA_topj}

Stage III aggregates candidate prompt embeddings via temperature-scaled soft fusion
(Eq.~\eqref{eq:fusion}).
In our implementation, we set $J$ to include \emph{all} expanded prompt candidates for each class,
and rely on score-based weighting (and a light gating rule) to downweight borderline prompts rather than
hard-truncating the pool.
We also report a \texttt{Top}$_J$ sweep by explicitly restricting the fusion to the \texttt{Top}$_J$ candidates ranked by the concept scoring in Eq.~\eqref{eq:scoring}.
When $J{=}1$, Eq.~\eqref{eq:fusion} degenerates to selecting a single best prompt embedding.

\begin{table}[h]
\centering
\setlength{\tabcolsep}{5pt}
\caption{\textbf{\texttt{Top}$_J$ sensitivity in Stage III.}
$J{=}1$ denotes single-prompt selection without fusion, while \textbf{All} uses the full expanded prompt set.}
\label{tab:supp_topj}
\vspace{-0.15cm}
\resizebox{0.49\linewidth}{!}{
\begin{tabular}{c c c c c c >{\columncolor{green!5}}c}
\toprule
\textbf{$J$} & 1 & 2 & 4 & 8 & 16 & All \\
\midrule
\textbf{NS Avg. mIoU} & 60.6 & 61.5 & 63.6 & 64.5 & 64.5 & \textbf{64.7} \\
\textbf{RS Avg. mIoU} & 48.1 & 49.7 & 50.1 & 51.7 & \textbf{52.1} & \textbf{52.1} \\
\bottomrule
\end{tabular}
}
\vspace{-0.2cm}
\end{table}

Restricting fusion to a small $J$ reduces prompt-level variance and can prevent weak candidates from entering the mixture.
In contrast, using \textbf{All} maximizes linguistic coverage and lets the temperature-scaled weights suppress
low-scoring prompts automatically, which is the default setting in our main experiments.

\subsection{Concept Scoring Metric: Dice \vs IoU}
\label{sec:suppA_metric}

In Eq.~\eqref{eq:scoring}, the main paper uses \texttt{Dice} to score candidate prompts on representative supports.
We ablate the scoring metric by replacing \texttt{Dice} with \texttt{IoU} while keeping all other steps unchanged.
We report this comparison in Table~\ref{tab:supp_metric}. Dice is often numerically more stable on small or thin structures due to its symmetric overlap form, whereas IoU is stricter and can be more sensitive to boundary errors.

\begin{table}[h]
\centering
\setlength{\tabcolsep}{6pt}
\caption{\textbf{Concept scoring in Stage III: Dice \vs IoU (\cf Eq.~\eqref{eq:scoring}).}}
\label{tab:supp_metric}
\vspace{-0.15cm}
\resizebox{0.44\linewidth}{!}{
\begin{tabular}{l c c}
\toprule
\textbf{Scoring metric} & \textbf{NS Avg. mIoU} & \textbf{RS Avg. mIoU} \\
\midrule
\rowcolor{green!5}
\textbf{Dice (main)} & \textbf{64.7} & \textbf{52.1} \\
IoU & 62.9 & 51.0 \\
\bottomrule
\end{tabular}
}
\vspace{-0.2cm}
\end{table}

\newpage
\section{Concept Bank Construction Cost}
\label{sec:suppB}

We report the one-time offline cost to construct \cb{} on the target support split.
This cost is \emph{not} included in the inference efficiency comparison (Table~\ref{tab:ablation-speed}),
since \cb{} is built once per dataset and then reused for all inference runs.
We measure wall-clock time of Stages I-III (\S\ref{sec:cb_construction}) on the support split:
(\textit{i}) mask-pooled crop embeddings and prototype estimation (Stage I),
(\textit{ii}) representative support mining (Stage II),
(\textit{iii}) candidate scoring and fusion (Stage III).

\begin{table}[h]
\centering
\setlength{\tabcolsep}{3.5pt}
\caption{\textbf{One-time \cb{} construction time per dataset (target support split, \ie, training set).}}
\label{tab:supp_time}
\vspace{-0.15cm}
\resizebox{0.88\linewidth}{!}{
\begin{tabular}{l c c c c c c c c}
\toprule
\textbf{Dataset} & $|\mathcal{C}|$ & \#Imgs & $M$ & Crops & Stage I (min) & Stage II (min) & Stage III (min) & Total (min)\\
\midrule
V21~\cite{pascal_voc}   & 21  & 1,464 & 27 & 840 & 0.99 & 1.00  & 1.03 & 3.02 \\
PC60~\cite{pascal_context}  & 60  & 4,996  & 6 & 2,399 & 2.68 & 2.83  & 1.63 & 7.14  \\
COCO-O~\cite{coco} & 80  & 118,287 & 40 & 3,240 & 4.32 & 4.61  & 2.62 & 11.55  \\
V20~\cite{pascal_voc}   & 20  & 1,464 &  14 & 800 & 0.88 & 0.94  & 0.72 & 2.55  \\
PC59~\cite{pascal_context}  & 59  & 4,996 &  6 & 2,359 & 2.63 & 2.78  & 1.65 & 7.07  \\
COCO-S~\cite{stuff} & 171 & 118,287  & 6 & 6,840 & 8.43 & 8.89  & 4.80 & 22.12  \\
City~\cite{Cityscapes}  & 19  & 2,975  & 9 & 760 & 1.25 & 1.50  & 0.79 & 3.53  \\
ADE~\cite{ade}   & 150 & 20,210 & 6 & 5,996 & 6.84 & 7.49  & 4.23 & 18.56  \\
\midrule
LoveDA~\cite{loveda}   & 7  & 2,522  & 12 & 2,800 & 1.70 & 0.22  & 0.35 & 2.27  \\
Potsdam~\cite{isprs_potsdam}  & 6  & 3,456  & 4 & 2,400 & 1.33 & 0.73  & 0.66 & 2.72  \\
Vaihingen~\cite{isprs_vaihingen} & 6  & 344  & 4 & 1,025 & 0.57 & 0.66  & 0.64 & 1.87  \\
iSAID~\cite{isaid}    & 16 & 33,978  & 6 & 6,282 & 5.51 & 0.64  & 0.42 & 6.56  \\
\bottomrule
\end{tabular}
}
\vspace{-0.2cm}
\end{table}

Stage I is dominated by visual feature extraction over masked crops and scales with the total number of crop pixels.
Stage III scales with $\sum_c |\mathcal{R}_c|\cdot M$ forward evaluations of $f_{\Phi}$ on representative supports.
Both stages are embarrassingly parallel across classes and supports, and benefit from batching and mixed precision.

\newpage
\section{Pseudo-code for Concept Bank Construction}
\label{sec:suppC}

{\small
\begin{verbatim}
# =========================================================
# ConceptBank Construction
# =========================================================
# Inputs:
#   D_train : target support set with images and GT masks
#   C       : class set
#   Phi     : frozen SAM3; provides f_Phi (mask predictor), phi_T (text encoder),
#             and the dense visual features used in mask pooling
#   Docs    : dataset documentation / label definitions (text)
# Hyper-params:
#   K : Top-K for representative mining
#   M : #candidate prompts per class (prompt expansion size)
#   J : Top-J for fusion
#   tau : softmax temperature for fusion

def build_concept_bank(D_train, C, Phi, Docs, K, M, J, tau):
    B = {}  # concept bank: class -> calibrated embedding e*_c
    # --- Stage I: Prototype Estimation ---
    # Collect mask-pooled crop embeddings per class
    Z = {c: [] for c in C}  # list of z(v,y) for each class
    for (x, Y) in D_train:
        for c in C:
            if has_instance(Y[c]):
                for (v, y_crop) in extract_instance_crops(x, Y[c]):
                    z = mask_pooled_embedding(Phi, v, y_crop)  # Eq.(6)
                    Z[c].append(z)
    P = {}
    for c in C:
        P[c] = l2_normalize(mean(Z[c]))  # prototype p_c, Eq.(7)

    # --- Stage II: Representative Support Mining (Top-K) ---
    R = {}
    for c in C:
        scores = [cos(z, P[c]) for z in Z[c]]
        idx = topk_indices(scores, k=min(K, len(scores)))
        R[c] = idx  # store indices of representative supports in Z[c]

    # --- Stage III: Prototype-consistent Concept Fusion ---
    for c in C:
        # Concept Pooling (prompt expansion from docs; text-only rewriting)
        Tc = prompt_expand(class_name=c, docs=Docs, M=M)  # list of strings
        Ec = [Phi.phi_T(t) for t in Tc]  # prompt embeddings

        # Concept Scoring (functional scoring via segmentation on R_c)
        S = []
        for m, e_cm in enumerate(Ec):
            dices = []
            for idx in R[c]:
                v, y = fetch_crop_and_mask(D_train, c, idx)
                y_hat = Phi.f_Phi(v, e_cm)
                dices.append(dice(y_hat, y))  # Eq.(10); or IoU in ablation
            S.append(mean(dices))

        # Top-J + soft fusion
        Jc = topj_indices(S, j=min(J, len(S)))
        w = softmax([S[j]/tau for j in Jc])
        e_star = sum(wi * l2_normalize(Ec[j]) for wi, j in zip(w, Jc))
        B[c] = e_star

    return B
\end{verbatim}
}

\section{Prompt Expansion Protocol and Generation Prompts}
\label{sec:suppD}

We expand each class name into a set of synonym- and attribute-enriched prompts using an off-the-shelf LLM.
The LLM is provided with dataset documentation (original paper, label taxonomy, and official label definitions).
The generation is constrained to \emph{text-level rewriting} of the given label semantics and does not introduce new visual concepts beyond the dataset definition.

\subsection{Quality-control rules}
\label{sec:suppD_qc}

We apply the following constraints to ensure label-faithful prompt expansion:
(\textit{i}) preserve the original class semantics; avoid adding new objects/scenes;
(\textit{ii}) allow synonyms, spelling variants (US/UK), plural/singular, and short attribute phrases
    that are consistent with the label definition;
(\textit{iii}) remove overly broad hypernyms if they may cause class leakage (\eg, replacing a fine-grained class with ``animal'');
(\textit{iv}) deduplicate near-identical strings; (\textit{v}) keep a fixed budget $M$ per class.

\subsection{LLM prompt templates}
\label{sec:suppD_prompts}

\paragraph{Template (Natural-scene datasets).}
\begin{quote}\small
\textbf{System:} You are an expert dataset annotator for open-vocabulary segmentation.
You must follow label definitions strictly and avoid introducing new visual concepts.

\textbf{User:}
You are given (1) a dataset name, (2) a class name, and (3) official label documentation.
Generate a comma-separated list of prompt variants that stay within the same class semantics.

\emph{Input:}
Dataset: \{DATASET\}
Class name: \{CLASS\}
Label documentation (authoritative): \{DOCS\}

\emph{Rules:}
- Output format: a single line, comma-separated phrases, no numbering, no extra text.
- Include: synonyms, spelling variants, common subtypes, and short attribute phrases consistent with the documentation.
- Do NOT add any object that is not implied by the documentation.
- Avoid overly generic words that could overlap many classes (\eg, ``thing'', ``object'').
- Keep 2-16 items total (including the original class name).
- Prefer concise noun phrases (1-4 words).
\end{quote}

\paragraph{Template (Remote-sensing datasets).}
\begin{quote}\small
\textbf{System:} You are an expert dataset annotator for aerial/remote-sensing semantic segmentation.
Follow dataset label definitions. Keep prompts consistent with top-down imagery and surface appearance.

\textbf{User:}
Given the dataset label documentation, expand the class name into a comma-separated list of synonyms and
attribute-enriched phrases that are label-faithful for aerial imagery.

\emph{Input:}
Dataset: \{DATASET\}
Class name: \{CLASS\}
Label documentation (authoritative): \{DOCS\}

\emph{Rules:}
- Output a single comma-separated line; no extra commentary.
- Allow: synonyms, material/surface descriptors (\eg, asphalt road), and aerial terms (\eg, roof, canopy) \emph{only if}
  they are consistent with the class definition.
- Do NOT introduce new land-cover categories not present in the documentation.
- Keep 1-15 items total (including the original class name).
\end{quote}

\subsection{Prompt Expansion Statistics}
\label{sec:suppA_prompt_stats}

Table~\ref{tab:supp_prompt_stats} reports summary statistics of the expanded concept candidates per class. Natural-scene object categories typically admit more lexical variants, while several remote-sensing benchmarks remain comparatively compact.

\begin{table}[h]
\centering
\setlength{\tabcolsep}{2.6pt}
\caption{\textbf{Prompt expansion statistics.} Counts are per-class candidate concepts (incl.\ the original class name).}
\label{tab:supp_prompt_stats}
\vspace{-0.18cm}
\resizebox{0.85\linewidth}{!}{
\begin{tabular}{l c c c c c c c c | c c c c}
\toprule
 & \multicolumn{8}{c}{\textbf{Natural-Scene}} & \multicolumn{4}{c}{\textbf{Remote-Sensing}} \\
\cmidrule(lr){2-9}\cmidrule(lr){10-13}
\textbf{Stat.} & \textbf{V21} & \textbf{V20} & \textbf{PC60} & \textbf{PC59} & \textbf{COCO-O} & \textbf{COCO-S} & \textbf{City} & \textbf{ADE} & \textbf{LoveDA} & \textbf{Potsdam} & \textbf{Vaihingen} & \textbf{iSAID} \\
\midrule
$|\mathcal{C}|$ & 21 & 20 & 60 & 59 & 81 & 171 & 19 & 150 & 7 & 6 & 6 & 16 \\
\textbf{Min}  & 4  & 4  & 3  & 3  & 2  & 2   & 4  & 3   & 4 & 2 & 2 & 1 \\
\textbf{Max}  & 27 & 14 & 6  & 6  & 40 & 6   & 9  & 6   & 12 & 4 & 4 & 6 \\
\textbf{Avg.}  & 9.29 & 8.40 & 4.67 & 4.64 & 5.47 & 4.93 & 6.32 & 4.83 & 8.00 & 2.67 & 3.17 & 3.50 \\
\bottomrule
\end{tabular}
}
\vspace{-0.22cm}
\end{table}

\newpage
\section{Additional Quantitative Results}
\label{sec:supp_quant}

We further compare against a recent \texttt{SAM3}-based approach, SegEarth-OV3~\cite{segearth-ov3}, on both natural-scene (NS) benchmarks (Table.~\ref{tab:ovs-ns-supp}) and remote-sensing (RS) benchmarks (Table.~\ref{tab:ovs-rs-supp}). 
Across the two groups, \cb{} improves the average mIoU by +\up{4.1} on \textbf{NS} and +\up{3.7} on \textbf{RS}. 
We note that SegEarth-OV3 relies on manually rewritten category names; in our re-checks, even small edits to a subset of class names can noticeably degrade its performance, suggesting a higher sensitivity to prompt phrasing. 
It is also substantially slower at inference time, since it follows the standard \texttt{SAM3} text-encoding pipeline and requires running the text encoder and grounding repeatedly per class (and even more when multiple prompts are used). 
In contrast, our Concept Bank is constructed offline and stores the calibrated text representations, so inference \emph{bypasses the text encoder entirely} (see \S\ref{sec:ablation}) and scores all labels for each sample in a single forward pass, yielding a clear speed advantage (often by several to tens of times) while improving robustness under drift.

\begin{table}[ht]
\centering
\setlength{\tabcolsep}{1.0pt}
\caption{\textbf{Quantitative comparison of OVS on natural-scene datasets.} ``Parameter-free'' denotes methods without learnable parameters or gradient-based optimization. Metric values are mIoU (\%). \colorbox{Best}{\best{bold}} and \colorbox{Second}{\second{underlined}} indicate the best and second-best results, respectively.}
\label{tab:ovs-ns-supp}
\vspace{-0.2cm}
\resizebox{\linewidth}{!}{
\begin{tabular}{l c c ccc ccccc c}
\toprule
\multirow{2}{*}{\textbf{Method}} &
\multirow{2}{*}{\textbf{Pub.\,\&\,Year}} &
\multirow{2}{*}{\textbf{Backbone (Size)}} &
\multicolumn{3}{c}{\textbf{\textit{with background}}} &
\multicolumn{5}{c}{\textbf{\textit{without background}}} &
\multirow{2}{*}{\textbf{Avg.}} \\
\cmidrule(lr){4-6}\cmidrule(lr){7-11}
& & & V21 & PC60 & COCO-O & V20 & PC59 & COCO-S & City & ADE & \\
\midrule
\rowcolor{black!5}
\multicolumn{12}{l}{\textbf{\textit{Training-based}}}\\
GroupViT~\cite{groupvit}  & CVPR'22 & GroupViT (ViT-S/16) &
50.4 & 18.7 & 27.5 & 79.7 & 23.4 & 15.3 & 11.1 & 9.2  & 29.4 \\
TCL~\cite{tcl}            & CVPR'23 & CLIP &
51.2 & 24.3 & 30.4 & 77.5 & 30.3 & 19.6 & 23.1 & 14.9 & 33.9 \\
SegCLIP~\cite{segclip}    & ICML'23 & CLIP &
52.6 & 24.7 & 27.5 & - & - & - & - & - & - \\
CoDe~\cite{code}          & CVPR'24 & CLIP &
57.7 & 30.5 & 32.3 & -   & -   & 23.9 & 28.9 & 17.7 & -   \\
SAM-CLIP~\cite{sam-clip}  & CVPR'24 & CLIP + SAM (ViT-B/16) &
60.6 & 29.2 & -  & -   & -   & 31.5  & -  & 17.1 & -   \\
Talk2DINO~\cite{talk2dino} & ICCV'25 & CLIP + DINOv2$^\dagger$ (ViT-B/14) &
65.8 & 37.7 & 45.1 & 88.5 & 42.4 & 30.2 & 38.1 & 22.5 & 46.3 \\
\midrule
\rowcolor{black!5}
\multicolumn{12}{l}{\textbf{\textit{Parameter-free}}}\\
NACLIP$^\ast$~\cite{naclip}     & WACV'25 & CLIP &
58.9 & 32.2 & 33.2 & 79.7 & 35.2 & 23.3 & 35.5 & 17.4 & 39.4 \\
LPOSS~\cite{lposs}  & CVPR'25 & CLIP + DINO (ViT-B/16) &
61.1 & 34.6 & 33.4 & 78.8 & 37.8 & 25.9 & 37.3 & 21.8 & 41.3 \\
CASS$^\ast$~\cite{cass}   & CVPR'25 & CLIP + DINO (ViT-B/8) &
65.8 & 36.7 & 37.8 & 87.8 & 40.2 & 26.7 & 39.4 & 20.4 & 44.4 \\
$\mathcal{DIH}$-CLIP~\cite{dih-clip}  & ICCV'25 & CLIP &
64.2 & 36.0 & 37.4 & 84.9 & 39.7 & 26.7 & 40.2 & 19.6 & 43.6 \\
SFP~\cite{sfp}~\cite{sfp}     & ICCV'25 & CLIP &
63.9 & 37.2 & 37.9 & 84.5 & 39.9 & 26.4 & 41.1 & 20.8 & 44.0 \\
CorrCLIP~\cite{corrclip}          & ICCV'25 & CLIP (ViT-L/14) + DINO (ViT-B/8) + SAM2 (Hiera-L) &
76.7 & 44.9 & 49.4 & 91.5 & 50.8 & 34.0 & 51.1 & 30.7 & 53.6 \\
Trident~\cite{trident}           & ICCV'25 & CLIP + DINO (ViT-B/16) + SAM (ViT-B/16) &
67.1 & 38.6 & 41.1 & 84.5 & 42.2 & 28.3 & 42.9 & 21.9 & 45.8 \\
ReME~\cite{reme}              & ICCV'25 & CLIP + DINOv2 (ViT-L/14) + SAM (ViT-L/16) &
82.2 & 44.6 & 48.2 & 93.2 & 53.1 & 33.3 & 59.0 & 28.2 & 55.2 \\
RF-CLIP~\cite{rf-clip}           & AAAI'26 & CLIP &
67.2 & 37.9 & 39.1 & 87.0 & 41.4 & 27.5 & 43.0 & 21.0 & 45.5 \\
\midrule
SAM3$^\ast$~\cite{samv3} & ICLR'26 & SAM3 (PE-L+/14) &
\second{81.9} & 46.1 & 65.4 & 88.9 & 50.0 & 33.3 & 62.3 & 31.8 & 57.5 \\
SegEarth-OV3$^\ast$~\cite{segearth-ov3} & arXiv'25 & SAM3 (PE-L+/14) &
79.8 & \second{50.7} & \second{67.6} & \second{96.8} & \second{58.8} & \second{42.8} & \second{69.7} & \second{37.5} & \second{63.0} \\
\textbf{\cb{} (Ours)} & - & SAM3 (PE-L+/14) &
\best{87.1} & \best{56.5} & \best{67.9} &
\best{97.4} & \best{63.0} & \best{46.4} & \best{75.1} & \best{43.3} & \best{67.1} \\
\bottomrule
\end{tabular}
}
\parbox{\linewidth}{\footnotesize
~\emph{Notes:} Unless otherwise specified, CLIP uses ViT-B/16~\cite{ViT}. ``$\ast$" denotes results reproduced in our work. ``$\dagger$" indicates DINOv2 equipped with registers~\cite{dinov2-reg}. ``-" indicates results are not reported or applicable in the original paper.}
\end{table}

\begin{table}[h]
\centering
\setlength{\tabcolsep}{2.3pt}
\caption{\textbf{Quantitative comparison of OVS on remote-sensing datasets.}``Parameter-free'' denotes methods without learnable parameters or gradient-based optimization. Metric values are mIoU (\%). \colorbox{Best}{\best{bold}} and \colorbox{Second}{\second{underlined}} indicate the best and second-best results, respectively.}
\label{tab:ovs-rs-supp}
\vspace{-0.25cm}
\resizebox{\linewidth}{!}{
\begin{tabular}{l c c c c c c c}
\toprule
\multirow{2}{*}{\textbf{Method}} &
\multirow{2}{*}{\textbf{Pub.\,\&\,Year}} &
\multirow{2}{*}{\textbf{Backbone (Size)}} &
\multicolumn{4}{c}{\textbf{\textit{with background}}} &
\multirow{2}{*}{\textbf{Avg.}} \\
\cmidrule(lr){4-7}
& & & LoveDA & Potsdam & Vaihingen & iSAID & \\
\midrule
\rowcolor{black!5}
\multicolumn{8}{l}{\textbf{\textit{Training-based}}}\\
SAN~\cite{san}  & CVPR'23 & CLIP (ViT-L/14@336px) &
25.3 & 37.3 & 39.2 & 49.6 & 37.9 \\
Cat-Seg~\cite{cat-seg}  & CVPR'24 & CLIP (ViT-L/14@336px) &
28.6 & 35.8 & 42.3 & \second{53.3} & 40.0 \\
SkySense-O~\cite{skysense-o}  & CVPR'25 & CLIP (ViT-L/14@512px) &
38.3 & 54.1 & 51.6 & 43.9 & 47.0 \\
RSKT-Seg~\cite{rskt-seg}  & AAAI'26 & CLIP (ViT-L/14@336px) + RemoteCLIP$^\ddagger$ (ViT-B/16) + DINO (ViT-B/32) &
33.2  & 38.4 &  42.7 & \best{54.3}  & 42.2 \\
\midrule
\rowcolor{black!5}
\multicolumn{8}{l}{\textbf{\textit{Parameter-free}}}\\
CLIP~\cite{clip}         & ICML'21 & CLIP &
12.4 & 15.6 & 10.8 & 7.5 & 11.6 \\
MaskCLIP~\cite{maskclip}   & ECCV'22 & CLIP &
27.8 & 33.9 & 29.9 & 14.5 & 26.5 \\
GEM~\cite{gem}           & CVPR'24 & CLIP&
31.6 & 39.1 & 36.4 & 17.7 & 31.2 \\
ProxyCLIP~\cite{ProxyCLIP} & ECCV'24 & CLIP + DINOv2$^\dagger$ (ViT-B/14) &
34.3 & 49.0 & 47.5 & 21.8 & 38.2 \\
SCLIP$^\ast$~\cite{sclip}       & ECCV'24 & CLIP &
30.4 & 39.6 & 35.9 & 16.1 & 30.5 \\
SegEarth-OV$^\ast$~\cite{segearth}     & CVPR'25 & CLIP &
36.9 & 48.5 & 40.0 & 21.7 & 36.8 \\
CorrCLIP$^\ast$~\cite{corrclip}          & ICCV'25 & CLIP + DINO (ViT-B/8) + SAM2 (Hiera-L) &
36.9 & 51.9 & 47.0 & 25.5 & 40.3 \\
\midrule
SAM3$^\ast$~\cite{samv3} & ICLR'26 & SAM3 (PE-L+/14) &
35.6 & 54.1 & 49.1 & 17.7 & 39.1 \\
SegEarth-OV3$^\ast$~\cite{segearth-ov3} & arXiv'25 & SAM3 (PE-L+/14) &
\second{47.4} & \second{57.8} & \second{60.8} & 27.6 & \second{48.4} \\
\textbf{\cb{} (Ours)} & - & SAM3 (PE-L+/14) &
\best{49.4} & \best{60.5} & \best{63.0} & 35.4 & \best{52.1} \\
\bottomrule
\end{tabular}
}
\parbox{\linewidth}{\footnotesize
~\emph{Notes:} Unless otherwise specified, CLIP uses ViT-B/16~\cite{ViT}. ``$\ast$" denotes results reproduced in our work. ``$\dagger$" indicates DINOv2 equipped with registers. ``$\ddagger$" indicates RemoteCLIP~\cite{remoteclip}, a CLIP-based remote sensing foundation model. }
\vspace{-0.35cm}
\end{table}

\newpage
\section{Additional Qualitative Results}
\label{sec:supp_vis}

\begin{figure*}[h]
    \centering
    \includegraphics[width=1.0\linewidth]{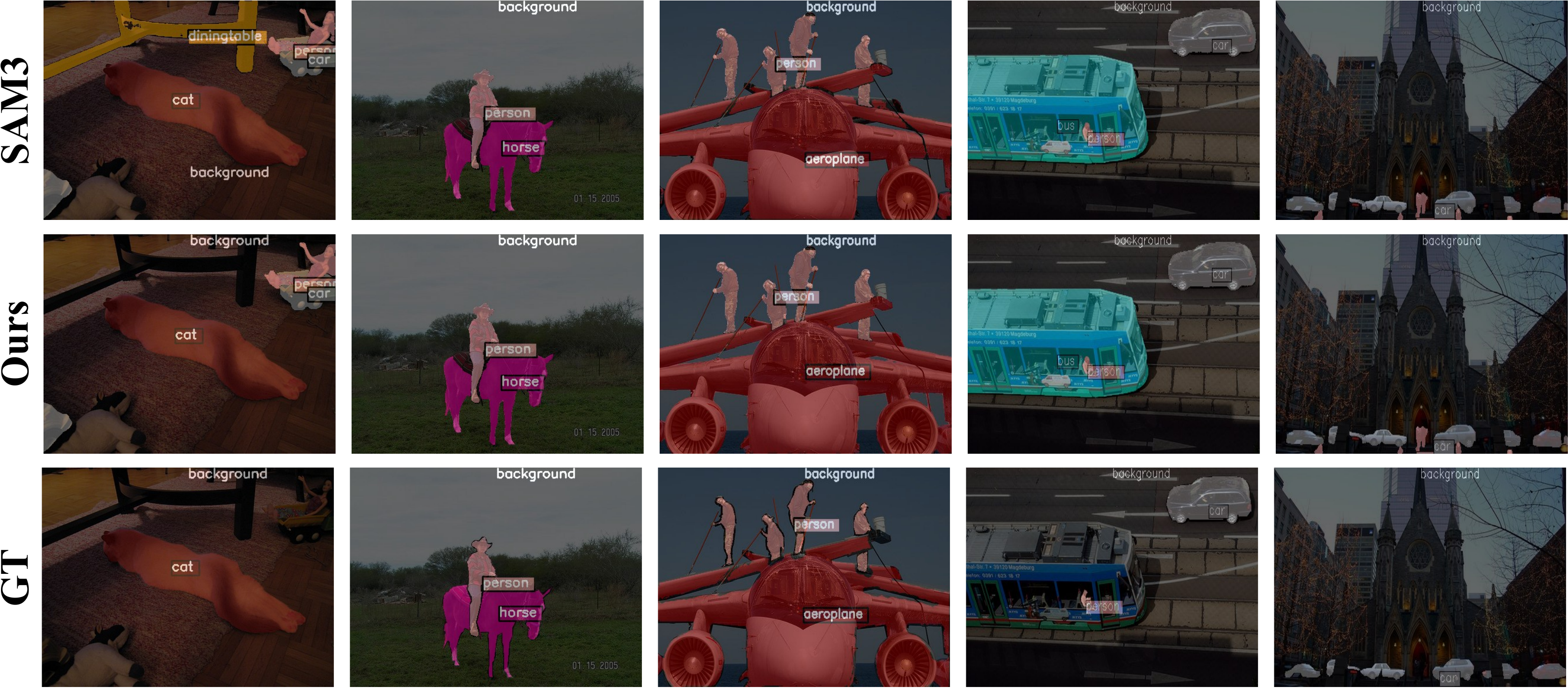}
    \caption{\textbf{Qualitative comparison results of open-vocabulary segmentation on Pascal VOC21.} Zoom in for best view.}
    \label{fig:supp_vis_voc21}
\end{figure*}

\begin{figure*}[h]
    \centering
    \includegraphics[width=1.0\linewidth]{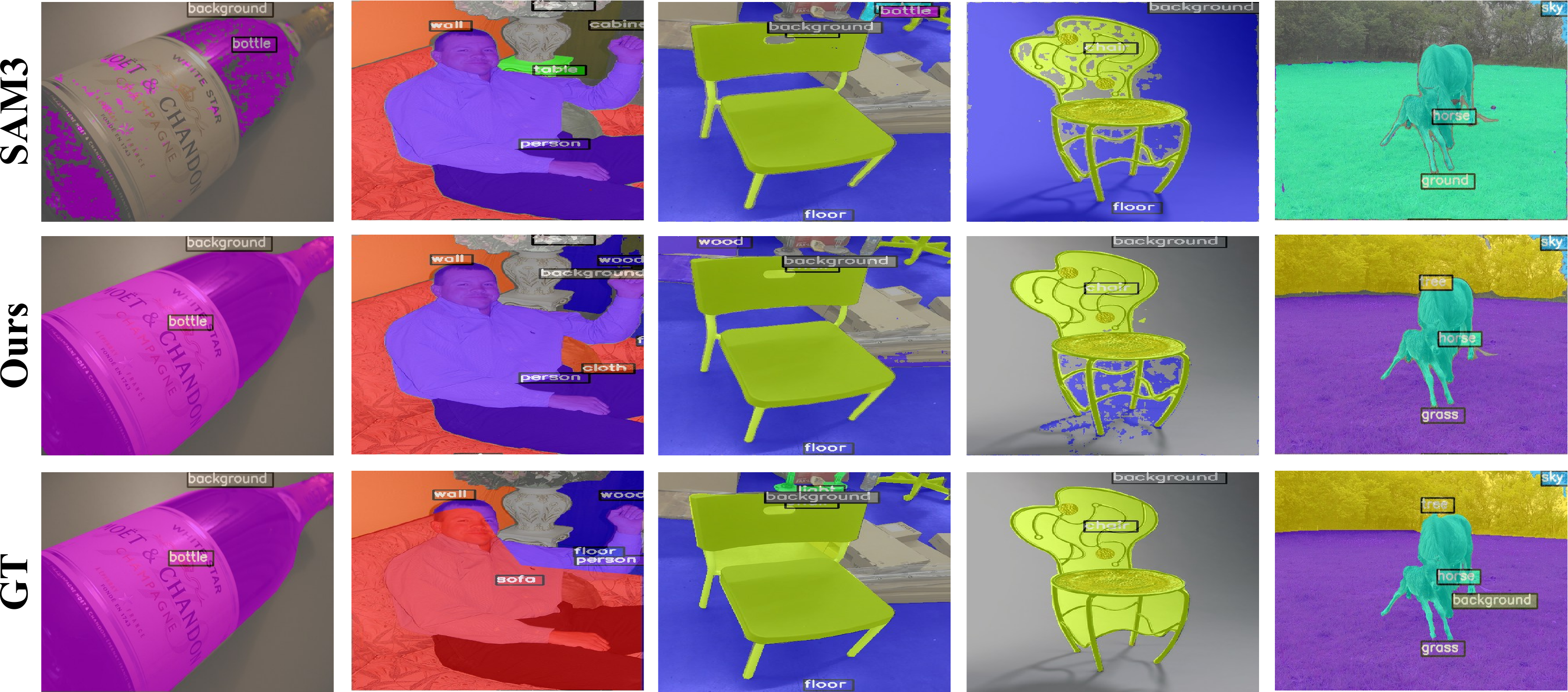}
    \caption{\textbf{Qualitative comparison results of open-vocabulary segmentation on Pascal Context 60.} Zoom in for best view.}
    \label{fig:supp_vis_ctx60}
\end{figure*}

\begin{figure*}[h]
    \centering
    \includegraphics[width=1.0\linewidth]{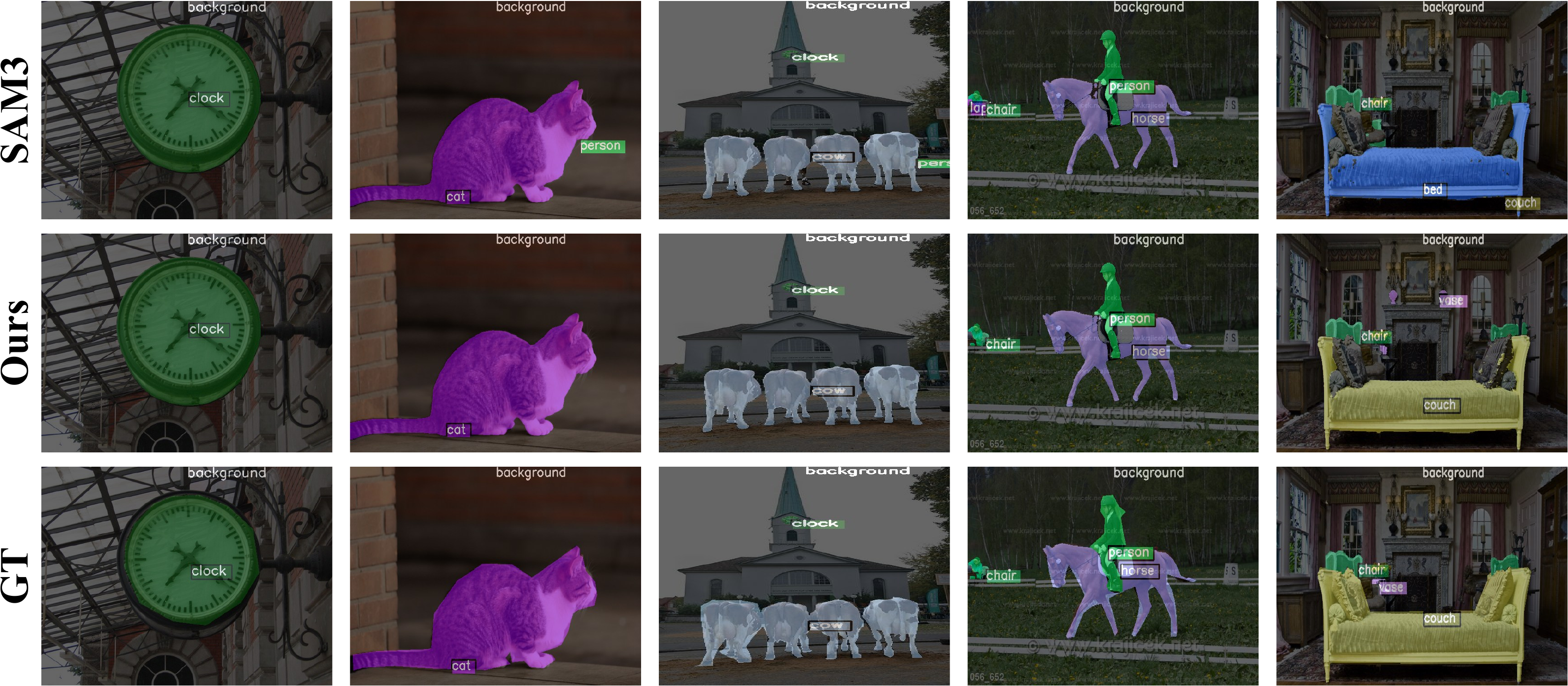}
    \vspace{-0.12cm}
    \caption{\textbf{Qualitative comparison results of open-vocabulary segmentation on COCO-Object.} Zoom in for best view.}
    \label{fig:supp_vis_coco_object}
    \vspace{-0.25cm}
\end{figure*}

\begin{figure*}[t]
    \centering
    \includegraphics[width=1.0\linewidth]{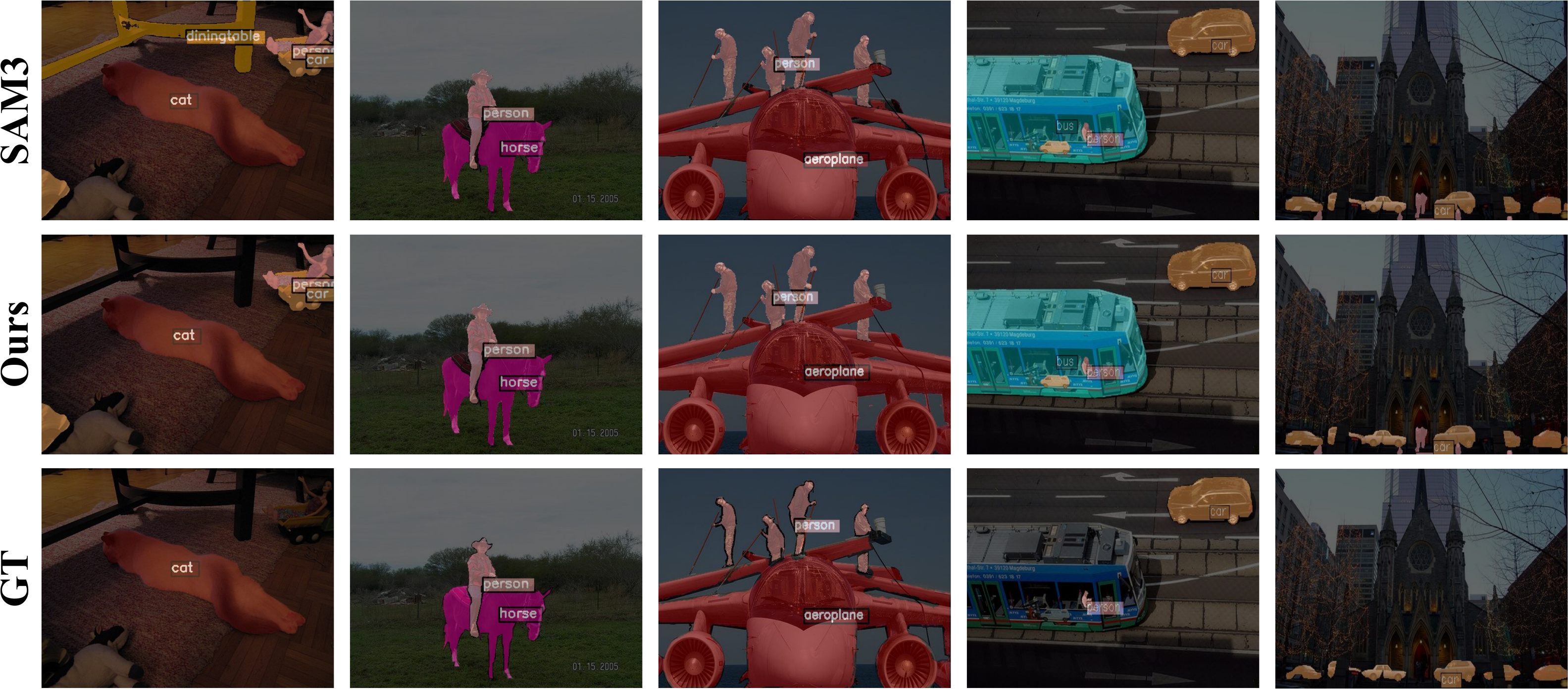}
    \vspace{-0.12cm}
    \caption{\textbf{Qualitative comparison results of open-vocabulary segmentation on Pascal VOC 20.} Zoom in for best view.}
    \label{fig:supp_vis_voc20}
    \vspace{-0.25cm}
\end{figure*}

\begin{figure*}[t]
    \centering
    \includegraphics[width=1.0\linewidth]{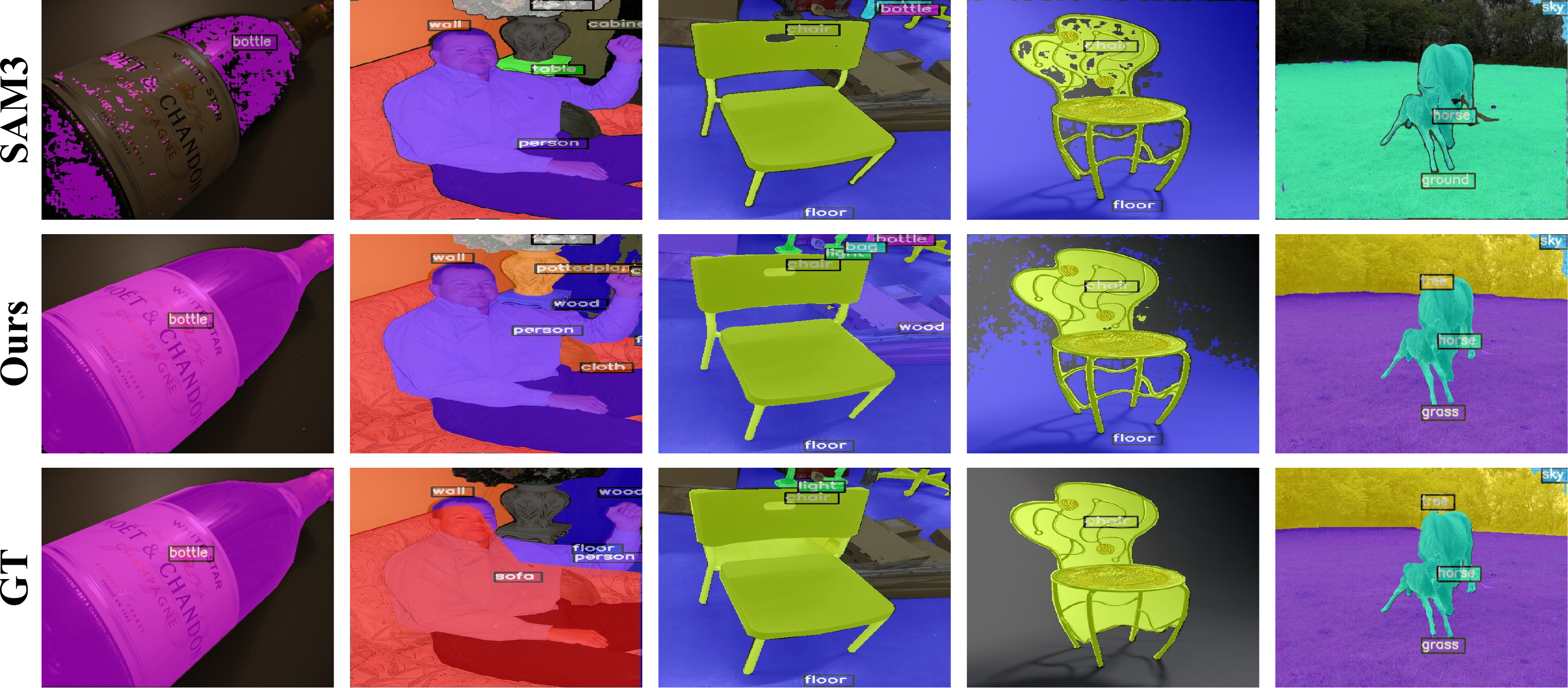}
    \vspace{-0.12cm}
    \caption{\textbf{Qualitative comparison results of open-vocabulary segmentation on Pascal Context 59.} Zoom in for best view.}
    \label{fig:supp_vis_ctx59}
    \vspace{-0.25cm}
\end{figure*}

\begin{figure*}[t]
    \centering
    \includegraphics[width=1.0\linewidth]{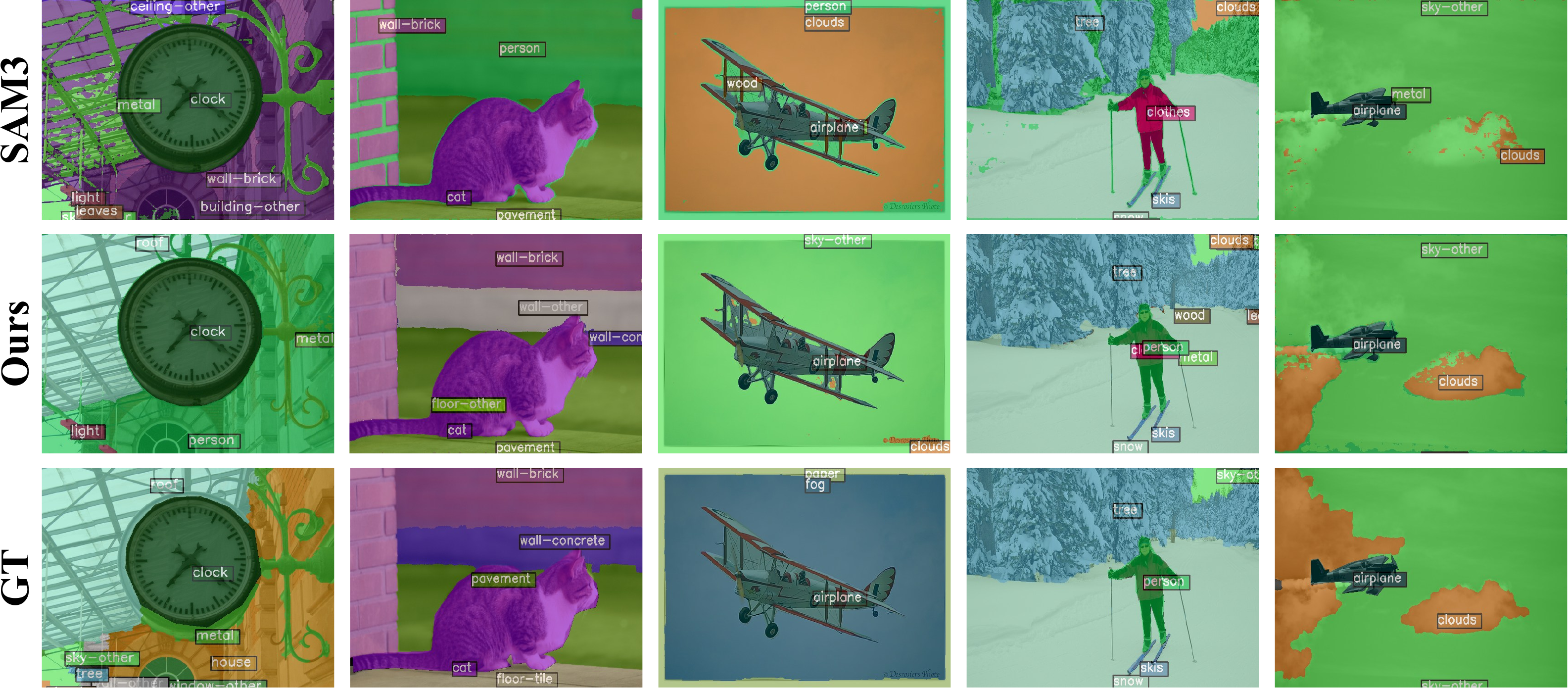}
    \vspace{-0.12cm}
    \caption{\textbf{Qualitative comparison results of open-vocabulary segmentation on COCO-Stuff.} Zoom in for best view.}
    \label{fig:supp_vis_coco_stuff}
    \vspace{-0.25cm}
\end{figure*}

\begin{figure*}[t]
    \centering
    \includegraphics[width=1.0\linewidth]{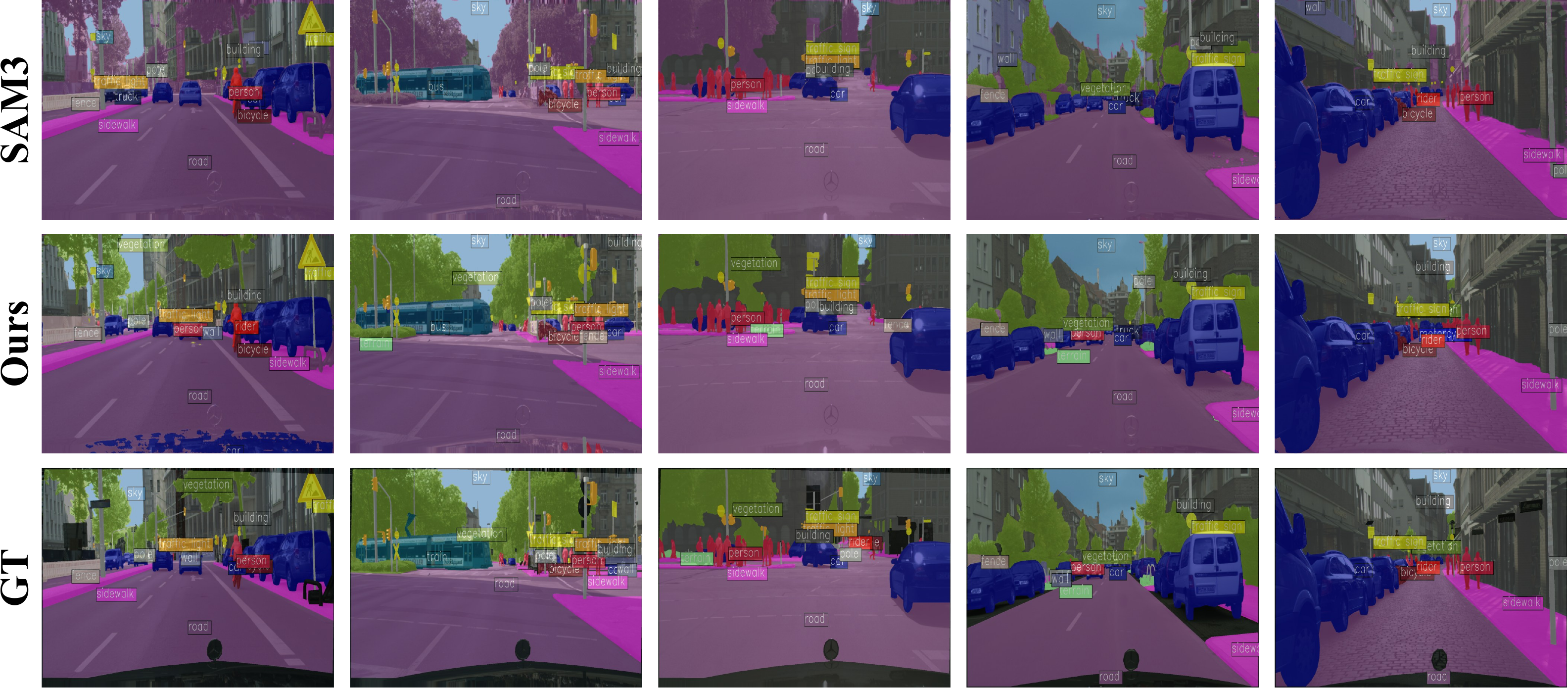}
    \vspace{-0.12cm}
    \caption{\textbf{Qualitative comparison results of open-vocabulary segmentation on Cityscapes.} Zoom in for best view.}
    \label{fig:supp_vis_cityscapes}
    \vspace{-0.25cm}
\end{figure*}

\begin{figure*}[t]
    \centering
    \includegraphics[width=1.0\linewidth]{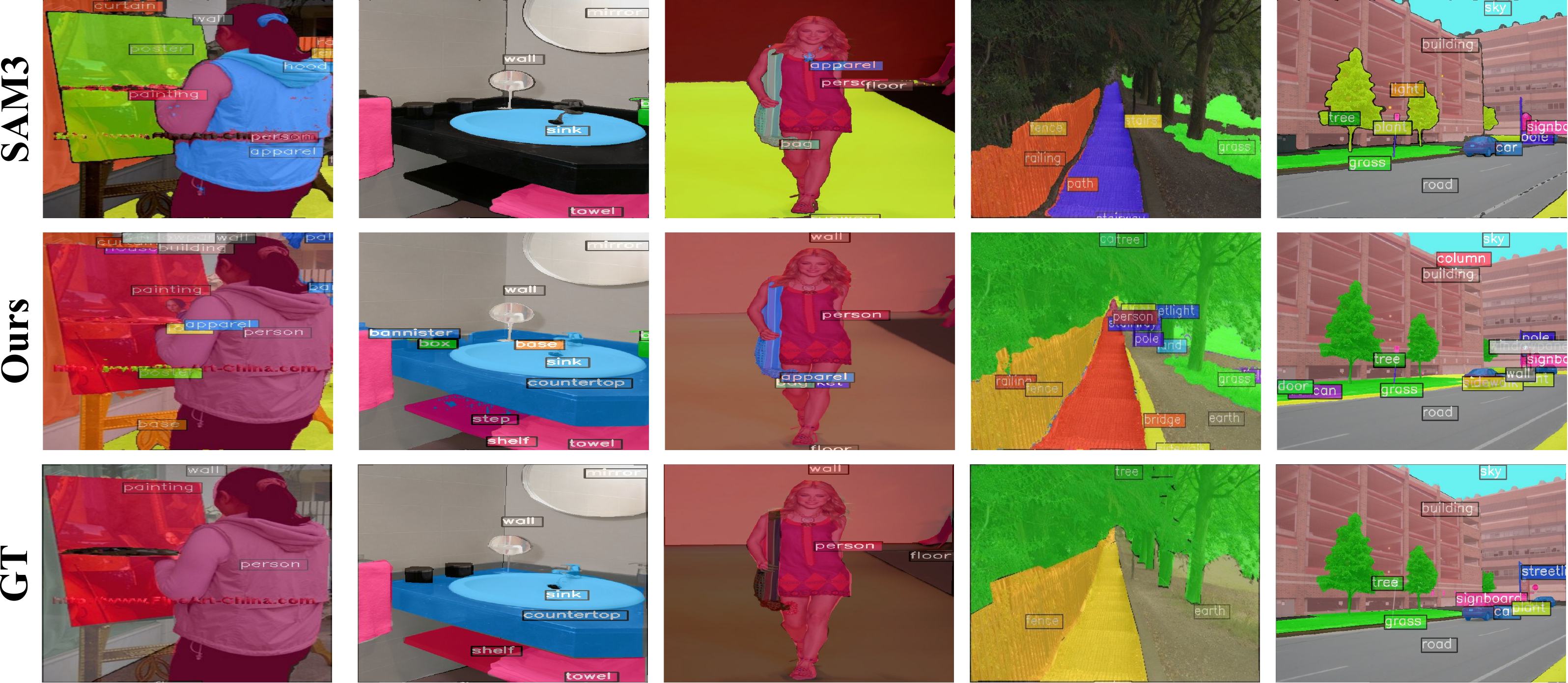}
    \vspace{-0.12cm}
    \caption{\textbf{Qualitative comparison results of open-vocabulary segmentation on ADE20K.} Zoom in for best view.}
    \label{fig:supp_vis_ade20k}
    \vspace{-0.25cm}
\end{figure*}

\begin{figure*}[h]
    \centering
    \vspace{-0.12cm}
    \includegraphics[width=1.0\linewidth]{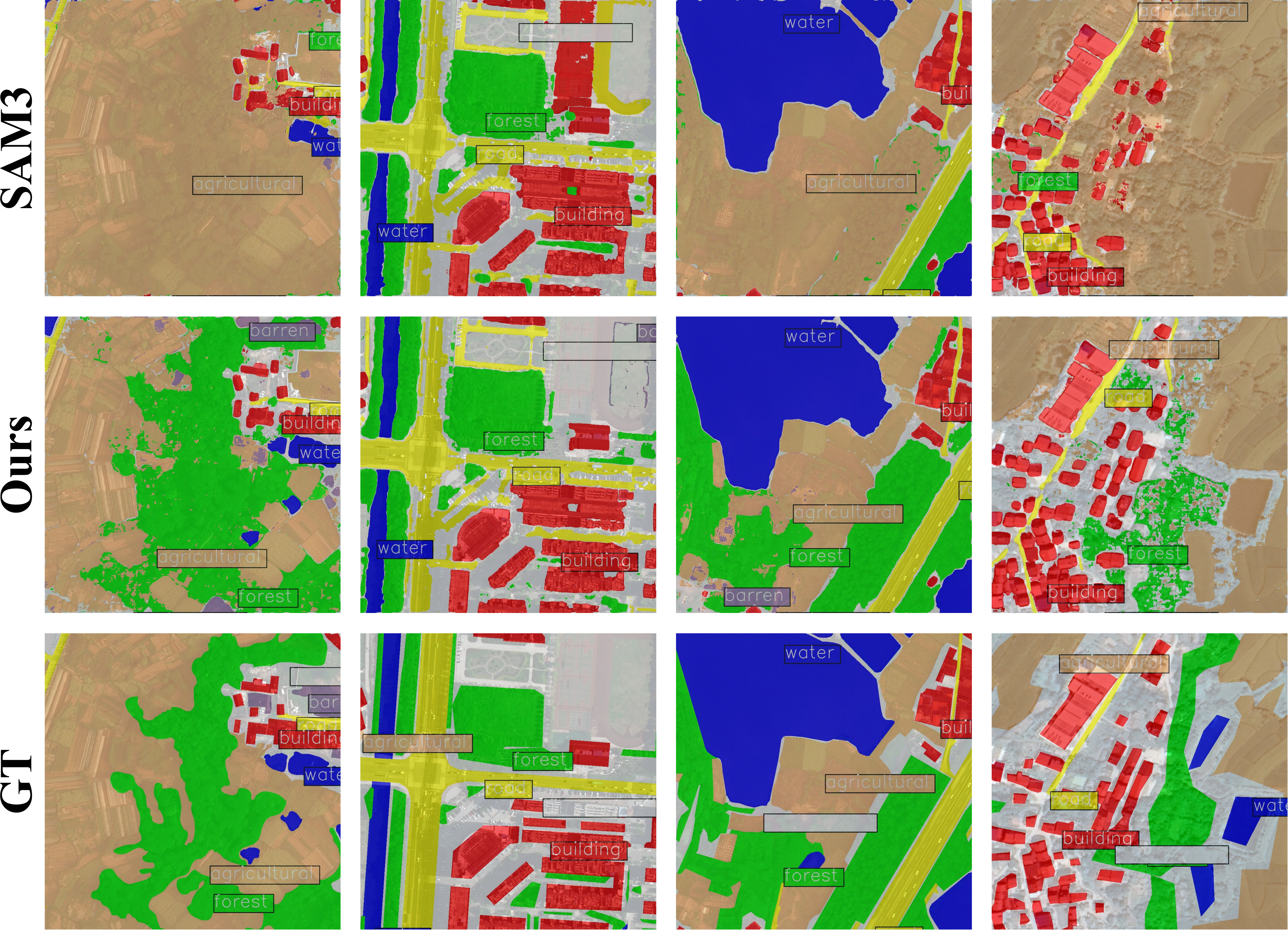}
    \caption{\textbf{Qualitative comparison results of open-vocabulary segmentation on LoveDA.} Zoom in for best view.}
    \label{fig:supp_vis_loveda}
    \vspace{-0.25cm}
\end{figure*}

\begin{figure*}[h]
    \centering
    \includegraphics[width=1.0\linewidth]{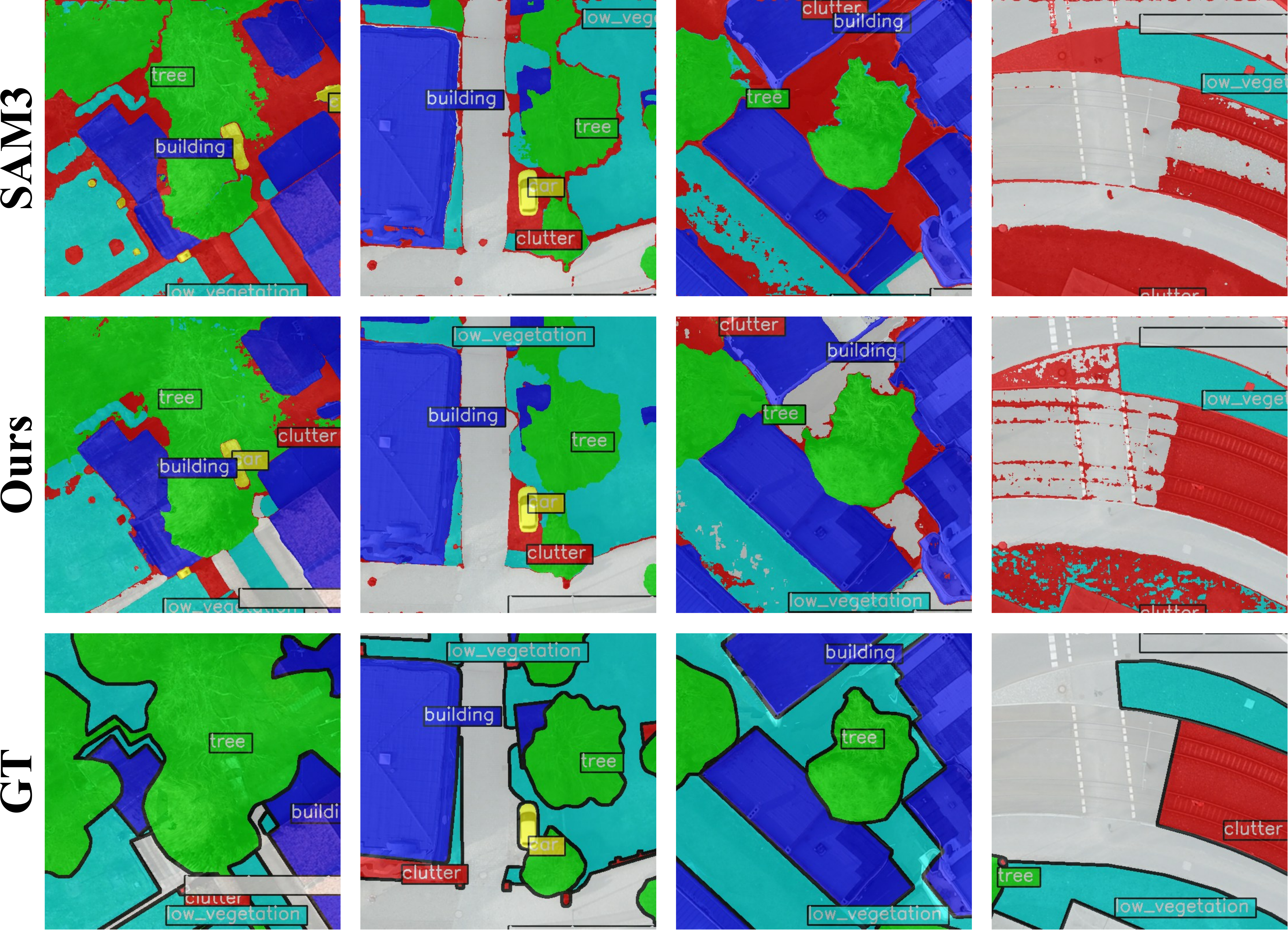}
    \vspace{-0.12cm}
    \caption{\textbf{Qualitative comparison results of open-vocabulary segmentation on Potsdam.} Zoom in for best view.}
    \label{fig:supp_vis_potsdam}
    \vspace{-0.25cm}
\end{figure*}

\begin{figure*}[h]
    \centering
    \includegraphics[width=1.0\linewidth]{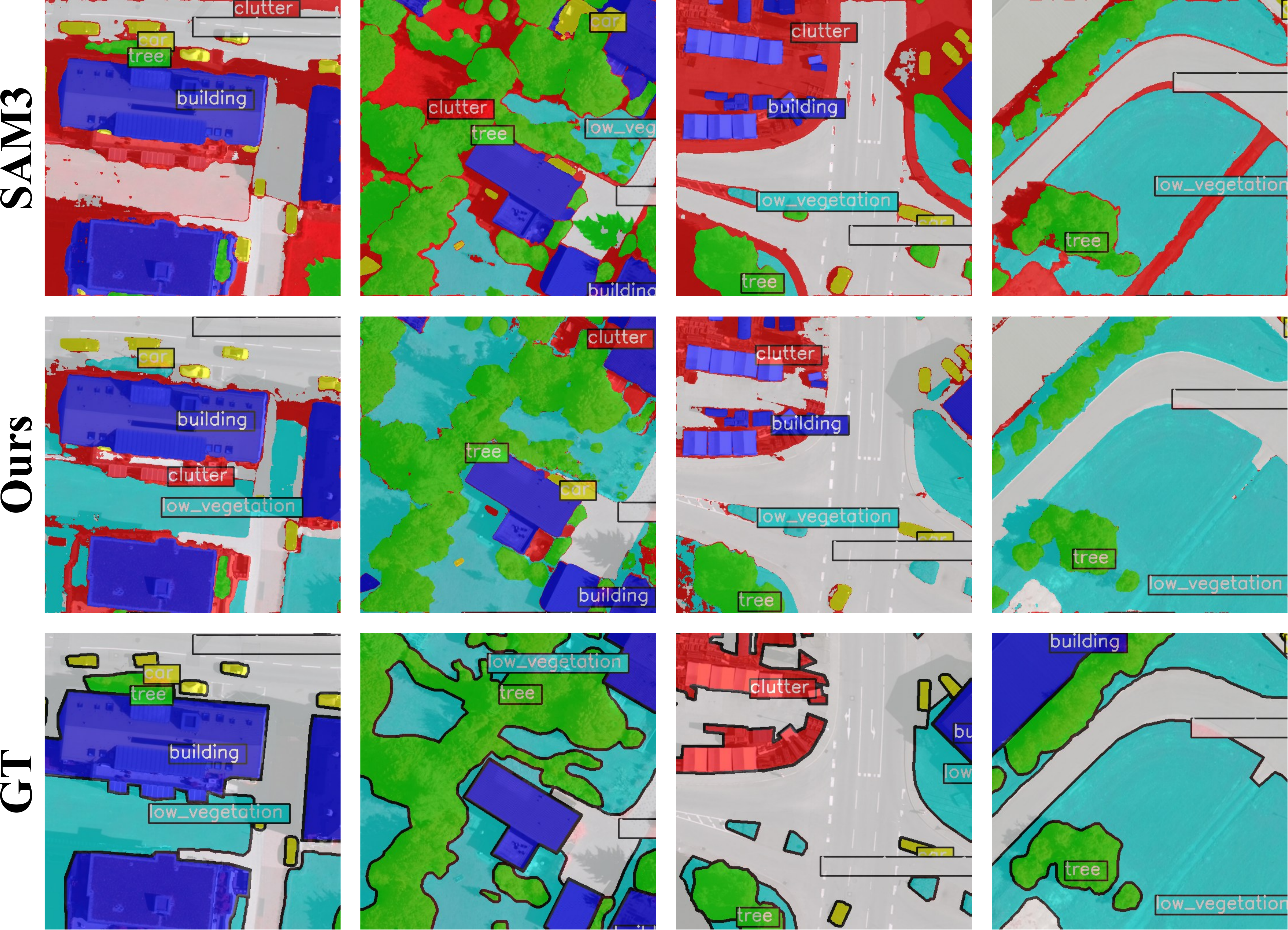}
    \vspace{-0.12cm}
    \caption{\textbf{Qualitative comparison results of open-vocabulary segmentation on Vaihingen.} Zoom in for best view.}
    \label{fig:supp_vis_vaihingen}
    \vspace{-0.25cm}
\end{figure*}

\begin{figure*}[h]
    \centering
    \includegraphics[width=1.0\linewidth]{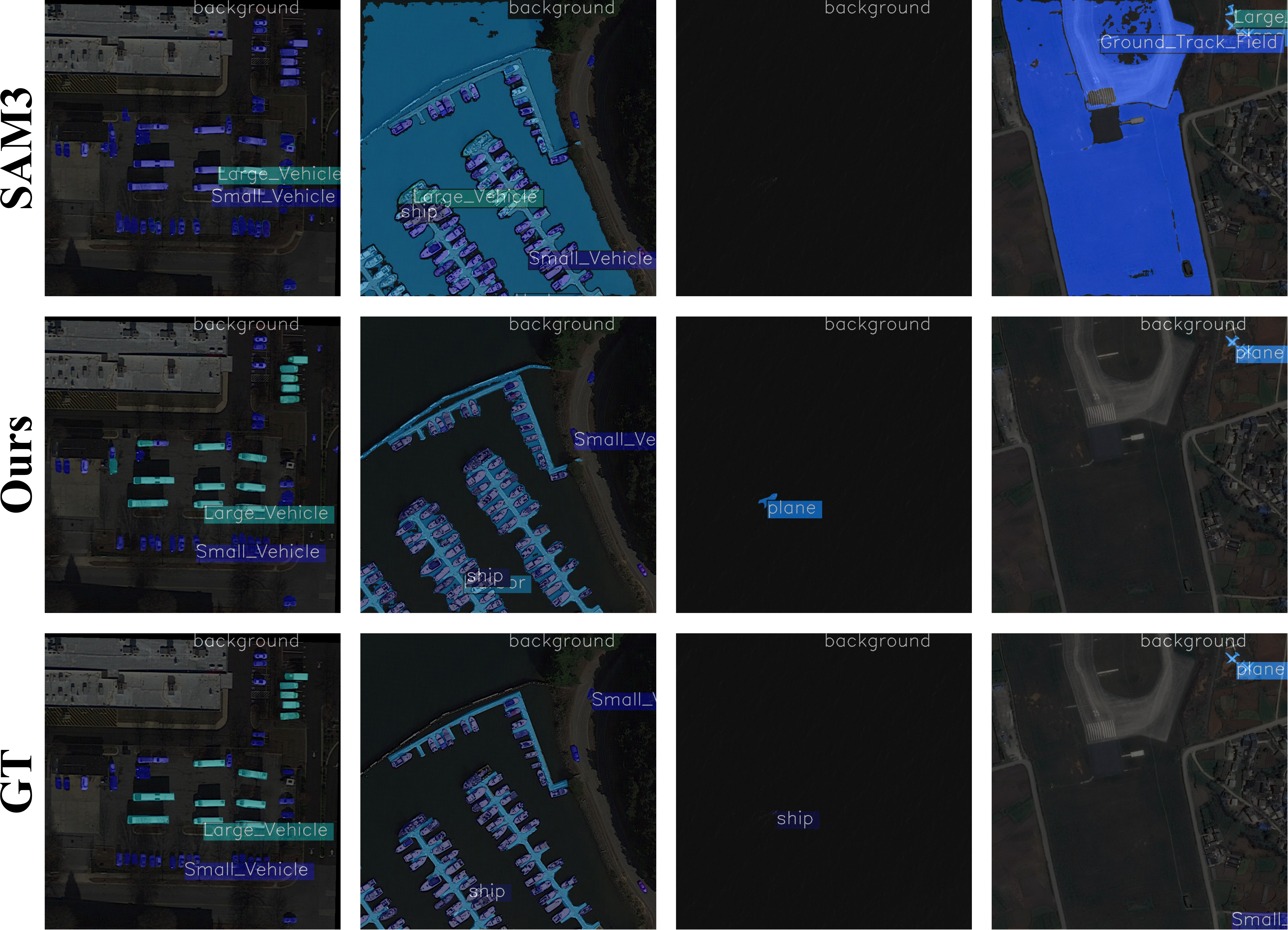}
    \vspace{-0.12cm}
    \caption{\textbf{Qualitative comparison results of open-vocabulary segmentation on iSAID.} Zoom in for best view.}
    \label{fig:supp_vis_isaid}
    \vspace{-0.25cm}
\end{figure*}

We present additional qualitative comparisons between the original \texttt{SAM3} baseline and our \cb{} on both Natural-Scene (see Figs.~\ref{fig:supp_vis_voc21}-\ref{fig:supp_vis_ade20k}) and Remote-Sensing (see Figs.~\ref{fig:supp_vis_loveda}-\ref{fig:supp_vis_isaid}) benchmarks. 
As shown in Figs.~\ref{fig:supp_vis_voc21}-\ref{fig:supp_vis_isaid}, \cb{} effectively reduces spurious activations caused by distribution shifts.
We also observe cases in which both methods produce visually plausible segmentations that are more consistent with object boundaries than the provided annotations.

\end{document}